\documentclass[10pt,twocolumn,letterpaper]{article}
\PassOptionsToPackage{table,dvipsnames}{xcolor}

\usepackage[pagenumbers]{cvpr}

\usepackage{makecell}
\usepackage[export]{adjustbox}
\usepackage{multirow}
\usepackage{array}
\usepackage{longtable}
\usepackage{float}
\usepackage{placeins}
\usepackage{algorithm}
\usepackage{algpseudocode}
\usepackage{tcolorbox}
\definecolor{cvprblue}{rgb}{0.21,0.49,0.74}
\usepackage[breaklinks,colorlinks,allcolors=cvprblue]{hyperref}

\def\paperID{}
\def\confName{CVPR}
\def\confYear{2025}

\title{\adjustbox{valign=c}{\includegraphics[height=1.4em]{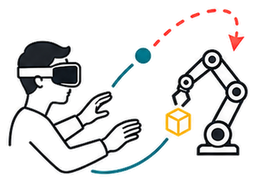}}\hspace{0.25em}EgoGenesis: Egocentric World-Action Modeling with \\
Online Anchored Projective Memory and Action-3D RoPE}
\author{
{\fontsize{10}{12}\selectfont
Zexuan Yan$^{1,2}$ \quad
Yuzhou Wu$^{3}$ \quad
Yue Ma$^{4}$ \quad
Zonghang He$^{1}$} \\[-0.15em]
{\fontsize{10}{12}\selectfont
Kaibo Yin$^{7}$ \quad
Xiaobing Tu$^{2}$ \quad
Yinggui Wang$^{2}$ \quad
Jinkui Ren$^{2}$} \\[-0.15em]
{\fontsize{10}{12}\selectfont
Xiantao Zhang$^{2}$ \quad
Shijian Wang$^{5}$ \quad
Jinghong Liu$^{6}$ \quad
Linfeng Zhang$^{1\dagger}$} \\[0.3em]
{\fontsize{10}{12}\selectfont
$^{1}$Shanghai Jiao Tong University \quad
$^{2}$Alibaba Group} \\[-0.1em]
{\fontsize{10}{12}\selectfont
$^{3}$Tianji KernalMind Co., Ltd. \quad
$^{4}$The Hong Kong University of Science and Technology} \\[-0.1em]
{\fontsize{10}{12}\selectfont
$^{5}$Southeast University \quad
$^{6}$Renmin University of China \quad
$^{7}$The University of Tokyo} \\[0.15em]
{\fontsize{10}{12}\selectfont
\textbf{Project: \href{https://egogenesis.github.io/}{\texttt{\textcolor{cyan}{https://egogenesis.github.io/}}}}}
}

\newcommand{\method}{\textsc{EgoGenesis}}
\newcommand{\egosim}{\textsc{EgoSim}}

\tcbset{
  promptbox/.style={
    colback=gray!4,
    colframe=black!65,
    boxrule=0.5pt,
    arc=1mm,
    left=6pt,
    right=6pt,
    top=5pt,
    bottom=5pt,
    fonttitle=\bfseries
  }
}

\makeatletter
\let\egogenesis@cvpr@maketitle\@maketitle
\renewcommand{\@maketitle}{%
  \egogenesis@cvpr@maketitle
  \vspace{-2.0em}
  \begin{center}
    \includegraphics[width=0.86\textwidth]{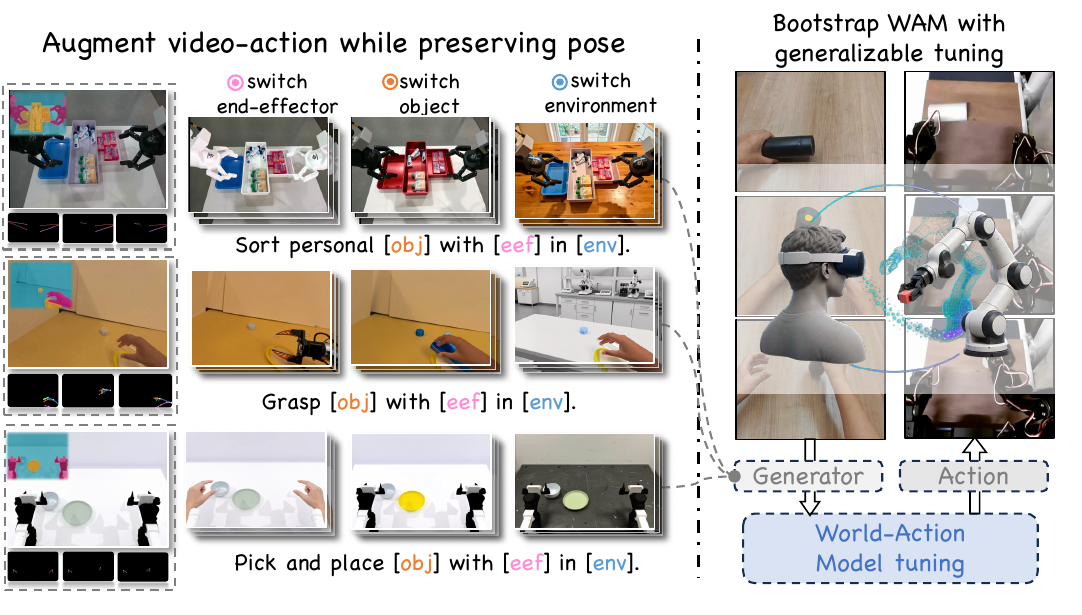}
    \captionof{figure}{\method{} expands scarce real demonstrations with controllable egocentric videos that improve downstream WAM generalization on real robots.}
    \label{fig:teaser}
  \end{center}
  \vspace{-0.7em}
}
\makeatother

\patchcmd{\abstract}{\vspace*{12pt}\noindent}{\vspace*{4pt}\noindent}{}{}

\begin{document}

\maketitle

{\renewcommand\thefootnote{\fnsymbol{footnote}}%
\footnotetext[2]{Corresponding author: Linfeng Zhang.}}

\begin{abstract}
Egocentric video offers rich manipulation experience for embodied AI, yet collecting diverse egocentric data across scenes, objects, motions, and embodiments remains costly. We present \method, an egocentric world-action simulator that synthesizes controllable, high-quality manipulation videos to expand scarce real-world training data. \method{} builds on a pretrained video generation prior and introduces two geometry-aware conditioning mechanisms. Online Anchored Projective Memory (OAPM) preserves a first-frame 3D scene anchor while periodically refreshing a recent state during autoregressive generation. Action-3D Rotary Position Embedding (A3D-RoPE) encodes end-effector motion with camera-aware 3D rotary coordinates, injecting action geometry into skeleton-to-video cross-attention for precise control. Together, these components improve visual fidelity, geometric stability, and action alignment in long egocentric rollouts. Moreover, augmenting 400 real trajectories with 400 \method-generated trajectories improves out-of-distribution real-robot success from 77\% to 84\% on single-arm tasks and from 53\% to 70\% on dual-arm tasks, demonstrating that the synthesized data substantially improve downstream WAM generalization.
\end{abstract}

\addtocontents{toc}{\protect\setcounter{tocdepth}{-1}}
\section{Introduction}
\label{sec:introduction}
In the current field of embodied AI, training World Action Models has increasingly become a promising and popular paradigm. Training these models requires egocentric observations paired with action trajectories across a diverse range of scenes, objects, and embodiments~\citep{wu2024gr1,li2026egowam,hu2025videopredictionpolicy}. Collecting such data with real robots is expensive: each new trajectory requires physical execution, reset, and supervision, while failures can damage objects or hardware. Large egocentric corpora demonstrate the value of this perspective for capturing hand-object contact and transferable skills, yet diverse, geometry-calibrated video--action pairs remain scarce~\citep{grauman2022ego4d,grauman2024egoexo4d,hoque2026egodex,kareer2025egomimic}. Recent video generation models and egocentric data engines offer a promising alternative: starting from limited demonstrations, they can generate additional egocentric interactions with varied appearances and environments~\citep{blattmann2023stablevideo,chen2024videocrafter2,wan2025wan,li2024egogen,wang2024egovid5m,hao2026egosim}. These synthetic rollouts are useful for downstream training when their visual observations remain consistent with the associated action labels.

\begin{figure}[t]
    \centering
    \includegraphics[width=\columnwidth]{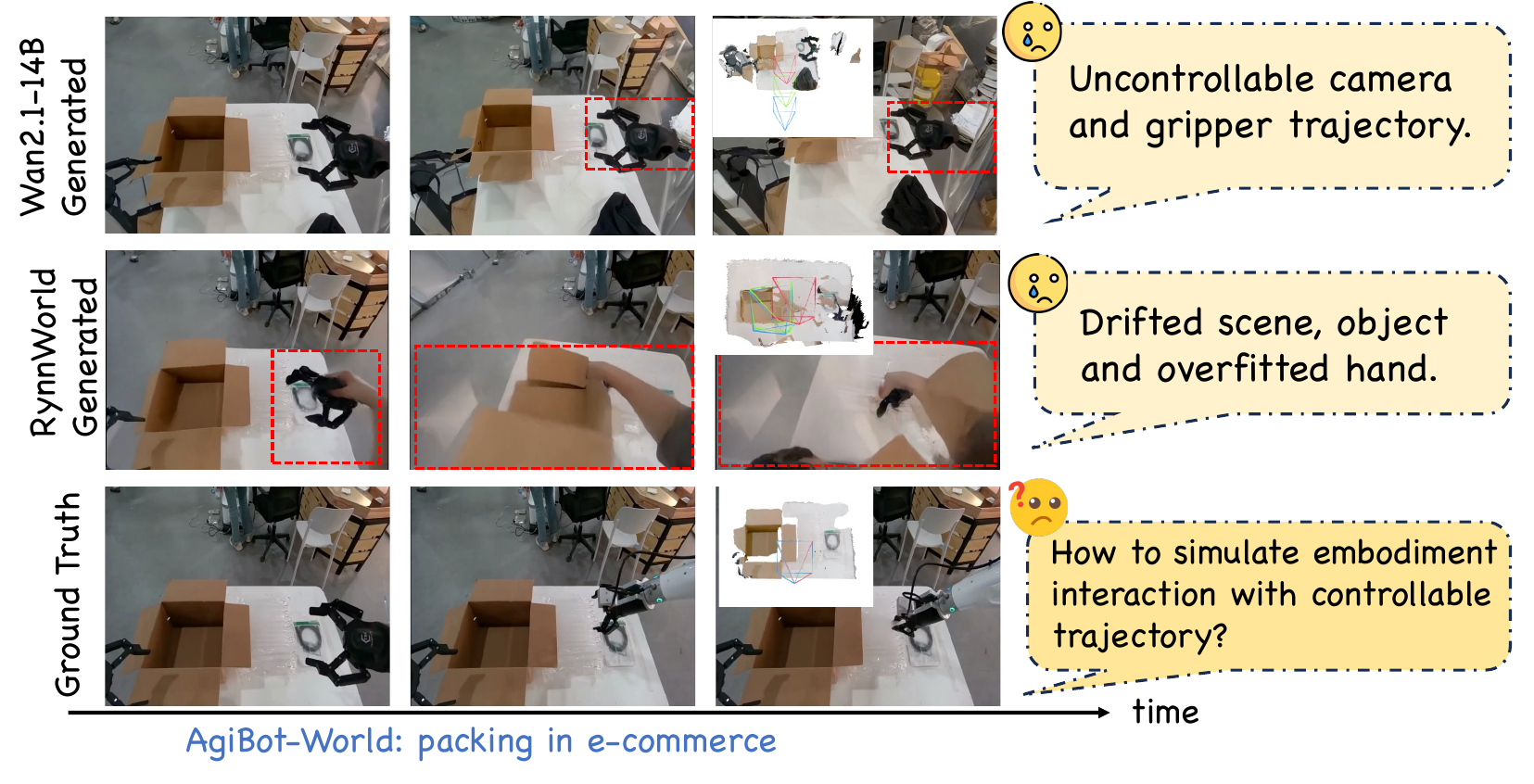}
    \caption{Motivation from diagnosing failures in existing egocentric video generators.}
    \label{fig:motivation}
\end{figure}

Existing egocentric video generators still struggle to simulate coherent, controllable interactions, as evidenced by the two representative failure modes in Figure~\ref{fig:motivation}. Wan2.1-14B~\citep{wan2025wan}, a strong general video prior, produces uncontrolled camera motion and a gripper trajectory that departs from the intended execution because neither trajectory is provided as an explicit geometric condition. Although controllable video methods introduce 2D skeletons, masks, point trajectories, or generic control images~\citep{wang2023videocomposer,yin2024dragnuwa,zhang2026sparse3dhand,chen2026handsonworld}, these signals do not fully constrain metric camera and end-effector motion. RynnWorld-TeleOp~\citep{zhao2026rynnworld}, by contrast, is action-conditioned, yet its generated scene and manipulated object drift over time, which we attribute to insufficient first-frame anchoring as occlusions and overfitted end-effector for training distribution~\citep{zhang2025egolcd}. These complementary failures motivate a central question: how can we simulate an embodied interaction with controllable camera and end-effector trajectories while preserving a stable scene and consistent object identity?

To address these geometric failures, we propose \method, an egocentric world-action model that generates controllable video--action pairs for embodied data augmentation, as summarized in Figure~\ref{fig:teaser}. \method{} innovatively involves decoupling the two modalities and processing them separately. For the world modeling, Online Anchored Projective Memory (OAPM) keeps an immutable first-frame 3D scene anchor while refreshing a compact, confidence-weighted recent snapshot during autoregressive generation. Camera-aware projective attention reads both states in the target view, allowing generated history to update the current scene without overwriting its stable anchor. For the action encoding, Action-3D Rotary Position Embedding (A3D-RoPE) rasterizes skeleton or end-effector joints onto the latent patch grid, unprojects them with camera trajectories , and encodes their reference-frame metric coordinates as rotary phases in skeleton-to-video cross-attention. Together, these components improve scene persistence and action alignment over long egocentric generation.

To improve the generalization of our approach and prevent overfitting to a specific scene or end-effector, we train \method{} on a source-balanced collection spanning human hands, dexterous hands, parallel grippers, and robot-arm end-effectors. The generated video--action pairs augment real demonstrations for tuning downstream world-action models and manipulation policies. On four bimanual and four single-arm real-robot skills, adding 400 generated trajectories to 400 real trajectories raises OOD success from 53\% to 70\% for bimanual tasks and from 77\% to 84\% for single-arm tasks (Table~\ref{tab:downstream_real} and Figure~\ref{fig:ood_stage_progress}). These gains demonstrate improved synthetic rollout quality from the use of \method{}: expanding scarce embodied data to improve real-world generalization.
\begin{figure*}[t]
    \centering
    \includegraphics[width=\textwidth]{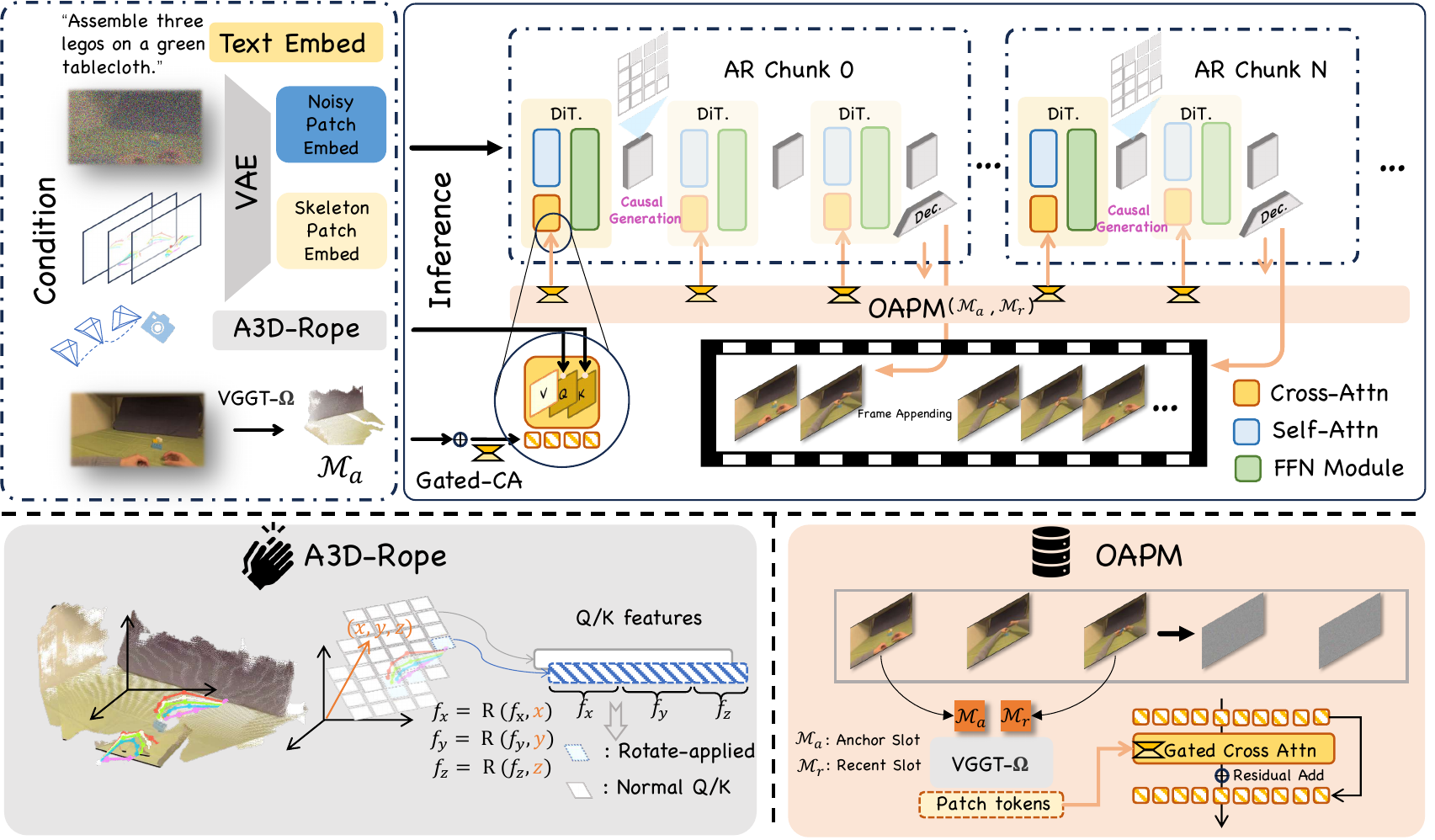}
    \caption{\method{} architecture. Text, noisy video, and skeleton embeddings provide the conditioning inputs. Autoregressive DiT generates and appends video frames. OAPM maintains anchored and recent 3D scene slots and refreshes the recent state online. A3D-RoPE injects metric action geometry into skeleton-to-video cross-attention.}
    \label{fig:pipeline}
\end{figure*}
In summary, our contributions are:
\begin{itemize}
    \item We introduce OAPM, an anchored projective 3D scene memory with online refresh for autoregressive egocentric generation.
    \item We introduce A3D-RoPE, which injects camera-aware metric skeleton and end-effector geometry into cross-attention for controllable action-aligned video generation.
    \item We show that augmenting real-robot data with \method-generated trajectories improves OOD generalization on bimanual and single-arm manipulation tasks.
\end{itemize}

\section{Related Work}
\label{sec:related_work}

\subsection{Video and World Action Models}
Large video diffusion and causal generators provide reusable priors for high-quality and long-horizon synthesis~\citep{blattmann2023stablevideo,chen2024videocrafter2,wan2025wan,nvidia2026cosmos3,zhu2026causal,li2026lingbotva}, while editing and control methods introduce masks, trajectories, camera motion, and entity-level conditions~\citep{geyer2023tokenflow,wang2023videocomposer,yin2024dragnuwa,wang2024motionctrl,akkerman2024interdyn,alzayer2026maskedvisualactions}. World-action models further use pixels, features, edited images, or imagined rollouts for prediction and robot control~\citep{yang2024unisim,wu2024ivideogpt,zhou2025dinowm,assran2025vjepa2,li2026egowam,zhang2026imagewam,guo2025ctrlworld,sun2026vlajepa,wu2024gr1,zhou2024robodreamer,hu2025videopredictionpolicy,du2025largevideoplanner}. Recent frontier video models including generic priors (Wan and Cosmos), interaction-conditioned generators (EgoHOI, Mask2IV, and CosHand), and egocentric models (\egosim{} and RynnWorld-Teleop)~\citep{li2026egohoi,li2026mask2iv,sudhakar2024coshand,hao2026egosim,zhao2026rynnworld} are unsatisfying in preventing geometric failures such as drifting. Besides, all of them treats end-effector simply without carefully modeling the geometric relationship between the end-effector and the scene in camera coordinate axes.

\subsection{Egocentric Interaction Data Generation}
Large egocentric corpora and human-to-robot pipelines provide egocentric hand--object observations, 3D geometry, and transferable action cues~\citep{grauman2022ego4d,grauman2024egoexo4d,hoque2026egodex,banerjee2025hot3d,shaw2024humanego,wang2026egoinfinity,kareer2025egomimic}. Synthetic data engines complement them by generating egocentric interactions from environments, video--action corpora, tasks, or explicit hand and camera controls~\citep{li2024egogen,wang2024egovid5m,zhao2025tasterob,zhang2025egolcd,zhang2026sparse3dhand,chen2026handsonworld,li2026egohoi,li2026mask2iv}. Camera and hand geometry can improve such control~\citep{li2025prope,wang2025vggt,pavlakos2024hamer}, while generated or semantically varied observations can improve policy generalization~\citep{bharadhwaj2024gen2act,chen2024semanticaugmentations,liu2023libero,mu2025robotwin,agibot2025world}. \method{} connects these directions by synthesizing action-aligned egocentric rollouts and using them to improve downstream WAM generalization with flexible switching between various environments, object manipulation, end effector observation attributes as synthetic video-action pairs.

\section{Preliminaries}
\label{sec:preliminaries}

\paragraph{Problem setup and latent video modeling.}
Given an egocentric manipulation sample, we observe an initial RGB frame $I_0$, a language instruction $y$, camera intrinsics and extrinsics $\{\mathcal{C}_b\}_{b=1}^{B}$ over $B$ temporal blocks, an initial scene representation, and the corresponding action geometry. The action may be a MANO hand skeleton, a dexterous-hand skeleton, or a robot end-effector/gripper trajectory in a unified keypoint format. Our goal is to generate a rollout $\widehat V$ that follows the specified action geometry while preserving the scene, camera motion, object identity, and plausible contact dynamics; $\widehat V$ can serve as both a world-model prediction and synthetic embodied data.

During training, let $V$ be the clean target video, $E_{\mathrm{vae}}$ and $D_{\mathrm{vae}}$ the frozen VAE encoder and decoder, and $Z=E_{\mathrm{vae}}(V)$ the video latent, partitioned into $B$ temporal blocks $Z=[Z_1,\ldots,Z_B]$. Let $\mathbf{c}_b$ collect the language, camera, scene-memory, and action conditions for block $b$. The causal key--value cache $\mathcal{K}_{<b}$ contains the committed blocks preceding $b$. At flow time $t\in[0,1]$, a clean block $Z_b$ is interpolated with Gaussian noise as
\begin{equation}
Z_b^t=(1-\sigma_t)Z_b+\sigma_t\varepsilon_b,
\qquad \varepsilon_b\sim\mathcal{N}(0,\mathbf{I}),
\label{eq:dit-forward}
\end{equation}
where $\mathbf{I}$ is the identity matrix and we use the linear schedule $\sigma_t=t$. The DiT predicts the flow field
\begin{equation}
\begin{aligned}
\widehat{v}_b^t&=F_\theta\!\left(Z_b^t,t;\mathbf{c}_b,\mathcal{K}_{<b}\right),\\
\mathcal{L}_{\mathrm{FM}}&=\mathbb{E}\left[\|\widehat{v}_b^t-(\varepsilon_b-Z_b)\|_2^2\right].
\end{aligned}
\label{eq:dit-objective}
\end{equation}
Sampling integrates the learned flow from noise to data with step size $\Delta t$, e.g. using the Euler update
\begin{equation}
Z_b^{t-\Delta t}=Z_b^t-\Delta t\,\widehat{v}_b^t.
\label{eq:dit-sampler}
\end{equation}
Once $Z_b^{0}$ is denoised, it is committed as $Z_b$ and appended to $\mathcal{K}_{<b+1}$; the next block is generated from this committed history and its block-aligned conditions.

\section{Method}

\label{sec:method}

\subsection{Model Overview}
\method{} instantiates the chunkwise DiT model of Section~\ref{sec:preliminaries} with three conditioning streams, as illustrated in Figure~\ref{fig:pipeline}. The first is the noisy video block $Z_b^t$, with its first frame pinned during generation. The second is scene memory: compact 3D patch tokens with metric coordinates and confidence. The third is action control, denoted by $\mathcal{S}_b$: a dense skeleton or end-effector latent with metric 3D coordinates. Text and image encoders provide task and appearance context. After flow integration reaches $t=0$, $Z_b^{0}$ is committed to the causal cache $\mathcal{K}_{<b+1}$ and may refresh the recent scene memory before block $b+1$.

\subsection{Online Anchored Projective Memory}
Static first-frame conditioning becomes stale as the camera moves and objects change state. OAPM therefore maintains two abstract scene slots: an immutable anchor $\mathcal{M}_a$ from the clean first frame and a replace-only recent slot $\mathcal{M}_r^b$ from generated history.
Before each block generation, we encode the two slots with the pretrained VGGT-$\Omega$ and directly use its 3D scene reconstruction features as the scene embedding:
\begin{equation}
\mathbf{M}_b=\operatorname{SceneEncode}_{\Omega}
\!\left(\mathcal{M}_a\oplus\mathcal{M}_r^b\right),
\label{eq:oapm-memory}
\end{equation}
where $\oplus$ denotes slot concatenation and $\operatorname{SceneEncode}_{\Omega}$ extracts the pretrained VGGT-$\Omega$ features. Selected DiT layers then use the resulting memory bank in a gated cross-attention update:
\begin{equation}
\begin{aligned}
Q=W_QH \quad K=W_K\mathbf{M}_b \quad V=W_V\mathbf{M}_b\\
H\leftarrow H+\operatorname{GatedCrossAttn}(Q,K,V).
\end{aligned}
\label{eq:oapm-read}
\end{equation}
This lets the hidden states read both the stable scene anchor and the current interaction state through the gated pathway in Figure~\ref{fig:pipeline}.

After every generated block, the result is decoded by the pretrained VAE and committed to the causal KV cache $\mathcal{K}_{<b+1}$. With refresh interval $s_r$, OAPM decodes the causally visible generated prefix, encodes its latest RGB frame into a new snapshot, and replaces only $\mathcal{M}_{r}^{b}$:
\begin{equation}
\begin{aligned}
\widehat I_b&=\operatorname{RecentFrame}\!\left(D_{\mathrm{vae}}(Z_{\le b})\right),\\
Z_{\le b}&=[Z_1,\ldots,Z_b],\\
\mathcal{M}_{r}^{b+1}&=
\begin{cases}
E_{3D}(\widehat I_b), & b\equiv0\pmod {s_r},\\
\mathcal{M}_{r}^{b}, & \text{otherwise.}
\end{cases}
\end{aligned}
\label{eq:oapm-refresh}
\end{equation}
The anchor is never overwritten since the initial frame provides the clearest and most stable conditions, and new generated frames are used to update the established portions of the scene memory by applying learned gating parameters. This update mechanism ensures scene consistency during the dynamic refreshing and generation process, effectively mitigating the issue of scene or object drifting.

\subsection{Action-3D Rotary Position Embedding}
\begin{figure*}[t]
    \centering
    \includegraphics[width=\textwidth]{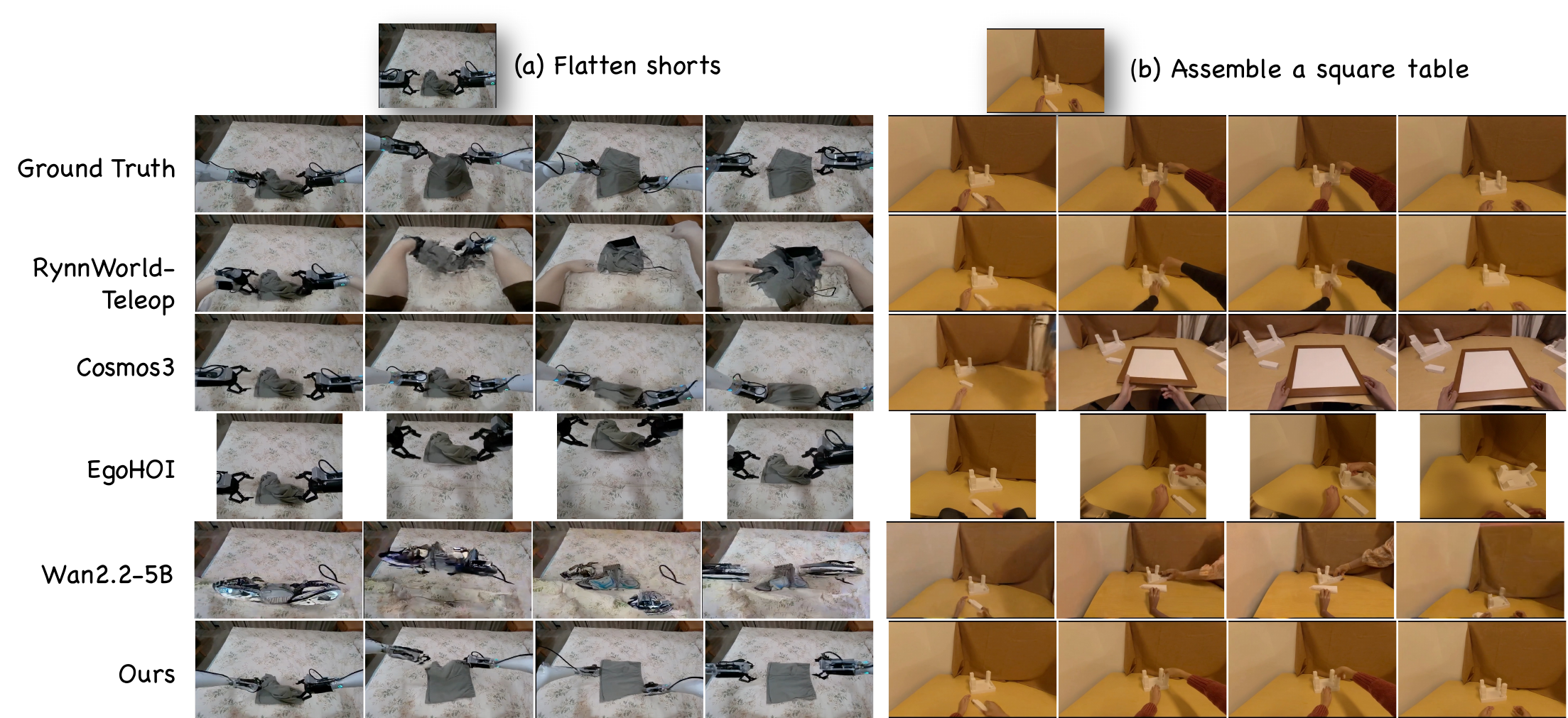}
    \caption{Qualitative comparison on flattening shorts and assembling a square table; \method{} better preserves embodiment, scene layout, and contact-driven object evolution.}
    \label{fig:qualitative_comparison}
\end{figure*}
\begin{table*}[t]
\centering
\scriptsize
\setlength{\tabcolsep}{4.2pt}
\renewcommand{\arraystretch}{1.08}
\resizebox{\textwidth}{!}{%
\begin{tabular}{>{\raggedright\arraybackslash}p{1.5cm}>{\raggedright\arraybackslash}p{3.9cm}ccccccc}
\toprule
Category &
Method &
\makecell{PSNR$\uparrow$} &
\makecell{SSIM$\uparrow$} &
\makecell{LPIPS$\downarrow$} &
\makecell{Kpt.Err$\downarrow$} &
\makecell{Phys.Faith$\uparrow$} &
\makecell{Subj. Cons.$\uparrow$} &
\makecell{Bg. Cons.$\uparrow$} \\
\midrule
\multirow{3}{*}{\makecell[l]{\textit{Generic video}\\\textit{models}}} & Wan2.1-14B-InP~\cite{wan2025wan} & 19.991 & 0.8288 & 0.3356 & 0.1295 & 0.7444 & 0.8786 & 0.9437 \\
& Wan2.2-5B-Control~\cite{wan2025wan} & 16.5896 & 0.7542 & 0.4323 & 0.0813 & 0.7296 & 0.7761 & 0.8956 \\
& Cosmos3-Nano~\cite{nvidia2026cosmos3} & 19.0996 & 0.8052 & 0.3810 & 0.1115 & \cellcolor{green!20}\underline{0.8093} & 0.8905 & 0.9392 \\
\midrule
\multirow{5}{*}{\makecell[l]{\textit{Egocentric}\\\textit{video models}}} & EgoHOI~\cite{li2026egohoi} & \cellcolor{green!20}\underline{20.223} & 0.7883 & 0.3326 & 0.1884 & 0.7537 & 0.8590 & 0.9193 \\
& Mask2IV$^{*}$~\cite{li2026mask2iv} & 18.6821 & 0.8102 & 0.3789 & 0.2086 & 0.6018 & 0.8786 & 0.9353 \\
& \egosim{}-14B~\cite{hao2026egosim} & 19.4114 & \cellcolor{green!20}\underline{0.8317} & \cellcolor{green!20}\underline{0.2750} & \cellcolor{green!20}\underline{0.0811} & 0.7759 & \cellcolor{blue!20}\textbf{0.9101} & \cellcolor{green!20}\underline{0.9491} \\
& CosHand~\cite{sudhakar2024coshand} & 18.7549 & 0.8033 & 0.4039 & 0.1202 & 0.4667 & 0.8089 & 0.9104 \\
& RynnWorld-TeleOp~\cite{zhao2026rynnworld} & 18.8247 & 0.8215 & 0.3913 & 0.2107 & 0.7870 & 0.8887 & 0.9466 \\
\midrule
\textit{Ours} & \textbf{\method{}} & \cellcolor{blue!20}\textbf{21.8609} & \cellcolor{blue!20}\textbf{0.8509} & \cellcolor{blue!20}\textbf{0.2399} & \cellcolor{blue!20}\textbf{0.0501} & \cellcolor{blue!20}\textbf{0.8278} & \cellcolor{green!20}\underline{0.8923} & \cellcolor{blue!20}\textbf{0.9546} \\
\bottomrule
\end{tabular}}
\caption{Generation quality on held-out trajectories under identical scene and action conditioning. Blue bold cells denote the best results, and green underlined cells denote the second-best results for each metric.}
\label{tab:main_results}
\end{table*}

Rendered skeleton videos provide visual control but do not expose metric depth. Our implementation keeps the pretrained WAN self-attention positional path unchanged and adds A3D-RoPE to the frame-local skeleton-to-video cross-attention adapters. The skeleton or end-effector latent is patchified on the video grid. Let $\mathcal{I}_b$ denote the patches covered by the rendered skeleton in block $b$; $\mathbf{X}_b$ collects their anchor-frame metric 3D coordinates, while $X_a$ denotes the scalar coordinate along axis $a\in\{x,y,z\}$.
\paragraph{Metric rotary cross-attention.}
At an A3D-RoPE adapter, queries originate from the video hidden states, whereas keys and values originate from the encoded skeleton tokens. In the following, $Q$, $K$, and $V$ refer only to the QKV entries indexed by $\mathcal{I}_b$ after their standard linear projections, rather than to QKV over the full patch grid. This limits metric rotations to patches with valid action coordinates and leaves background tokens unchanged. A3D-RoPE splits the selected query and key channels into three groups,
\begin{equation}
Q=[Q^x\,\|\,Q^y\,\|\,Q^z],
\qquad
K=[K^x\,\|\,K^y\,\|\,K^z].
\label{eq:a3d-split}
\end{equation}
Within each group, adjacent channels form standard two-dimensional RoPE pairs. If $M_a$ pairs are assigned to axis $a\in\{x,y,z\}$, pair $m$ uses the channel pair $(u,v)$:
\begin{equation}
\begin{aligned}
\theta_{a,m}(X_a)&=sX_a\kappa^{-m/M_a},\\
\begin{bmatrix}u'\\v'\end{bmatrix}
&=
\begin{bmatrix}
\cos\theta_{a,m}(X_a)&-\sin\theta_{a,m}(X_a)\\
\sin\theta_{a,m}(X_a)&\cos\theta_{a,m}(X_a)
\end{bmatrix}
\begin{bmatrix}u\\v\end{bmatrix},
\end{aligned}
\label{eq:a3d-rotate-simple}
\end{equation}
Here, $\theta_{a,m}(X_a)$ is the rotation angle induced by the metric coordinate $X_a$ at the $m$-th RoPE frequency, with $s=4$ and $\kappa=10^4$. For each supported patch in block $b$, its entry in $\mathbf{X}_b$ is represented by the anchor-frame coordinates $(X_x,X_y,X_z)$; thus, $X_a$ is one axis component of $\mathbf{X}_b$, rather than a separate coordinate. Applying Eq.~\eqref{eq:a3d-rotate-simple} along all three axes and to every supported patch yields the blockwise rotation $R_{\mathbf{X}_b}$, which is applied directly to $Q$ and $K$:
\begin{equation}
\widetilde Q=R_{\mathbf{X}_b}(Q),
\qquad
\widetilde K=R_{\mathbf{X}_b}(K).
\label{eq:a3d-attn-simple}
\end{equation}
The rotated $Q/K$ and selected $V$ update only $H_{\mathcal{I}_b}$ through the gated cross-attention pathway shown in Figure~\ref{fig:pipeline}. Coordinate construction and implementation details are provided in the supplementary material.

\section{Experiments}
\label{sec:experiments}

\subsection{Experimental Setup}
We evaluate \method{} as both a generative world-action model and a data augmentation engine on a mixed held-out benchmark containing test subsets conditioned on robot grippers, gripper end-effector (EEF) trajectories, and human-hand skeletons. All training and test splits are strictly disjoint, with no clip or trajectory overlap. Every video is standardized to 81 frames at 16 FPS and a spatial resolution of $832\times480$. Training uses a source-balanced 210K-clip corpus drawn from EgoDex~\citep{hoque2026egodex}, AgiBot~\citep{agibot2025world}, RoboTwin~\citep{mu2025robotwin}, Real-world Ego, and DexJoCo. In the first stage, we conducted 6k steps of SFT using pre-processed momory slots $\mathcal{M}_a$ and $\mathcal{M}_r$. For the second stage for autoregressive training, we utilized generated frame as $\mathcal{M}_r$ for auto-regressive training, which also ran for 6k steps.

We report PSNR, SSIM, and LPIPS for frame-level fidelity; Kpt.Err (Hand Keypoint End-Point Error) for action alignment; Phys.Faith, assessed by Kimi K2.7, for plausible contact and object motion; and Subj. Cons. (Subject Consistency) and Bg. Cons. (Background Consistency) for temporal stability. Kpt.Err is computed only on the 50-clip EgoDex subset. \method{} is trained on 8 NVIDIA A100 GPUs, while downstream real-robot inference and LingBot-VA training and testing are performed on 8 NVIDIA H100 GPUs. We compare with general image-to-video and controllable video baselines, including Wan2.1-Fun-14B-Inp, Wan2.2-5B-Control, EgoHOI, Mask2IV, and \egosim{}-14B. Further details are provided in the Supplementary Material.

\subsection{Qualitative Results}

Generalization across embodiments and tasks remains challenging, as evidenced by Figure~\ref{fig:qualitative_comparison}, where we compare the AgiBot-World shorts-flattening task and the EgoDex square-table assembly task. RynnWorld-TeleOp exhibits strong gripper overfitting, replacing the human hand with gripper-like morphology, while its assembled object also deviates substantially from the reference. Cosmos3 and EgoHOI follow the shorts-flattening instruction poorly and produce weak contact interactions; Cosmos3 additionally hallucinates an unintended object during table assembly. Wan2.2-5B-Control suffers severe drift and loses hand and end-effector cues. In contrast, \method{} remains close to the ground truth in both tasks while preserving coherent hand--object and gripper--object interactions.
\subsection{Quantitative Results}

To test whether geometry-aware conditioning improves interaction quality without sacrificing appearance, we compare \method{} with generic and egocentric generators under identical scene and action conditions. \method{} ranks first on six of seven metrics and second on subject consistency, as reported in Table~\ref{tab:main_results}. In particular, its Kpt.Err of 0.0501 and Phys.Faith of 0.8278 provide direct evidence of stronger action alignment and contact plausibility, while the best PSNR (21.8609), LPIPS (0.2399), and background consistency (0.9546) rule out a trade-off in visual fidelity or temporal stability. The joint gains therefore support more reliable action-conditioned interaction rather than appearance alone.

\subsection{Ablation Study}
To verify that the gains arise from both OAPM and A3D-RoPE, we vary one component at a time while fixing the Wan2.2-5B-Control backbone, training protocol, and complementary component. With A3D-RoPE fixed, adding the recent slot $\mathcal{M}_r$ to the anchor $\mathcal{M}_a$ raises PSNR from 20.4135 to 21.8609 and reduces LPIPS from 0.2533 to 0.2399, as shown in Table~\ref{tab:ablation_main}; this supports the value of tracking the evolving scene rather than relying only on the first frame. With OAPM fixed, A3D-RoPE outperforms both RoPE and PRoPE on all three compact fidelity metrics, including an SSIM increase from 0.8198 and 0.8408 to 0.8509. These controlled improvements attribute the full-model gains to both scene-state maintenance and metric action encoding. Furthermore, we tested a baseline AR model without any of our proposed methods. The results showed that it underperformed across all relevant metrics, demonstrating that each of our proposed techniques is individually effective.

\begin{table}[t]
    \centering
    \scriptsize
    \setlength{\tabcolsep}{1.8pt}
    \renewcommand{\arraystretch}{1.05}
    \resizebox{\columnwidth}{!}{%
    \begin{tabular}{@{}p{0.37\columnwidth}p{0.20\columnwidth}*{3}{>{\centering\arraybackslash}p{0.115\columnwidth}}@{}}
    \toprule
    Component Setting & Setting & PSNR$\uparrow$ & SSIM$\uparrow$ & LPIPS$\downarrow$ \\
    \midrule
    \textbf{Wan2.2-5B-Control-AR} & None & 19.9238 & 0.7812 & 0.3028 \\
    \midrule
    $\mathcal{M}_a$ only & \multirow{2}{*}{A3D-RoPE fixed} & 20.4135 & 0.8385 & 0.2533 \\
    +\textbf{$\mathcal{M}_r$ (OAPM)} & & \textbf{21.8609} & \textbf{0.8509} & \textbf{0.2399} \\
    \midrule
    RoPE~\cite{su2024roformer} & \multirow{3}{*}{OAPM fixed} & 21.4250 & 0.8198 & 0.2838 \\
    PRoPE~\cite{li2025prope} & & 21.8421 & 0.8408 & 0.2481 \\
    \textbf{A3D-RoPE} & & \textbf{21.8609} & \textbf{0.8509} & \textbf{0.2399} \\
    \bottomrule
    \end{tabular}}
    \caption{Compact core-component ablation on Wan2.2-5B-Control. The complete action, physical-faithfulness, and consistency metrics are reported in the supplementary material.}
    \label{tab:ablation_main}
    \end{table}

\begin{figure}[t]
    \centering
    \includegraphics[width=\columnwidth]{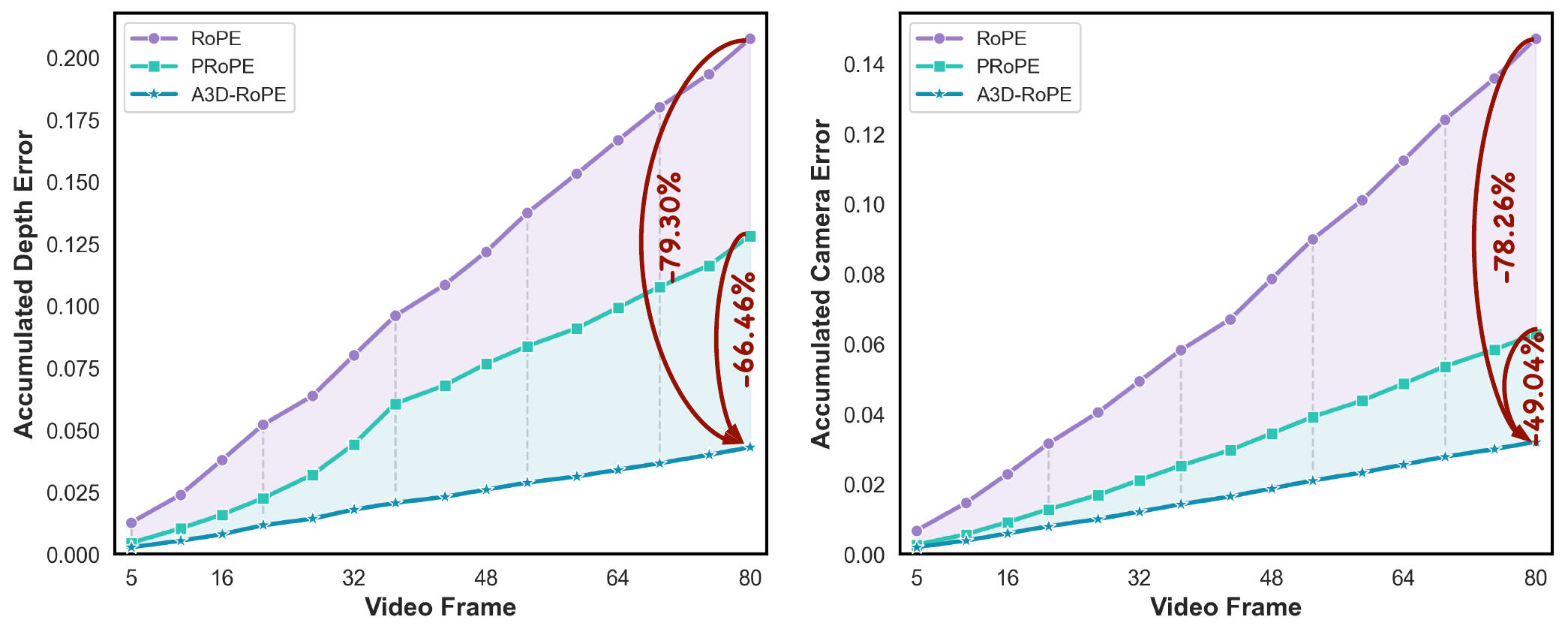}
    \caption{Accumulated depth and camera errors over an 80-frame rollout; A3D-RoPE consistently reduces geometric drift.}
    \label{fig:a3d_geometry_ablation}
\end{figure}

\paragraph{Geometric drift.}
A3D-RoPE substantially reduces long-horizon geometric drift, as demonstrated by the controlled comparison in Figure~\ref{fig:a3d_geometry_ablation}, which keeps OAPM and the training setup fixed. Depth-ERR is clip-scale-aligned AbsRel within action-keypoint regions, and Cam-ERR is the per-pixel $\ell_2$ error on first-camera-relative Pl\"ucker coordinates; both are computed with VGGT-$\Omega$ on 15 sampled non-anchor frames per clip. RoPE provides no metric action geometry, while PRoPE improves camera-aware video self-attention but does not encode the skeleton trajectory in metric 3D within action cross-attention. A3D-RoPE aligns the action keys and video queries with reference-frame 3D rotary coordinates, substantially reducing both local depth drift around the hand or end-effector and global camera-ray inconsistency. At frame 80, A3D-RoPE reduces Depth-ERR and Cam-ERR by 79.30\% and 78.26\% relative to RoPE, and by 66.46\% and 49.04\% relative to PRoPE, respectively. This sustained reduction shows that A3D-RoPE preserves action geometry throughout long autoregressive rollouts rather than improving only frame-level appearance; the corresponding Kpt.Err comparison is reported in the supplementary component-ablation table.

\begin{figure}[t]
    \centering
    \includegraphics[width=\columnwidth]{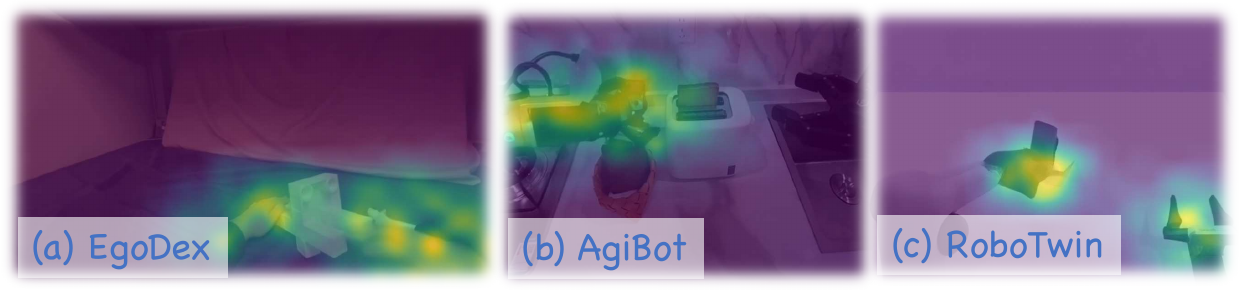}
    \caption{A3D-RoPE concentrates spatial influence on end-effectors and contact-relevant regions across embodiments.}
    \label{fig:a3d_attention_influence}
\end{figure}

\paragraph{Spatial influence of A3D-RoPE.}
A3D-RoPE concentrates its spatial influence on end-effectors and nearby contact regions, as visualized in Figure~\ref{fig:a3d_attention_influence}; warmer regions indicate larger changes in the cross-attention output when metric 3D encoding is enabled. By injecting end-effector information, object-interaction depth, and camera displacement relative to the hand frame, A3D-RoPE consistently focuses the response on interaction-relevant regions rather than the background across EgoDex, AgiBot, and RoboTwin.

\begin{figure}[t]
    \centering
    \includegraphics[width=\columnwidth]{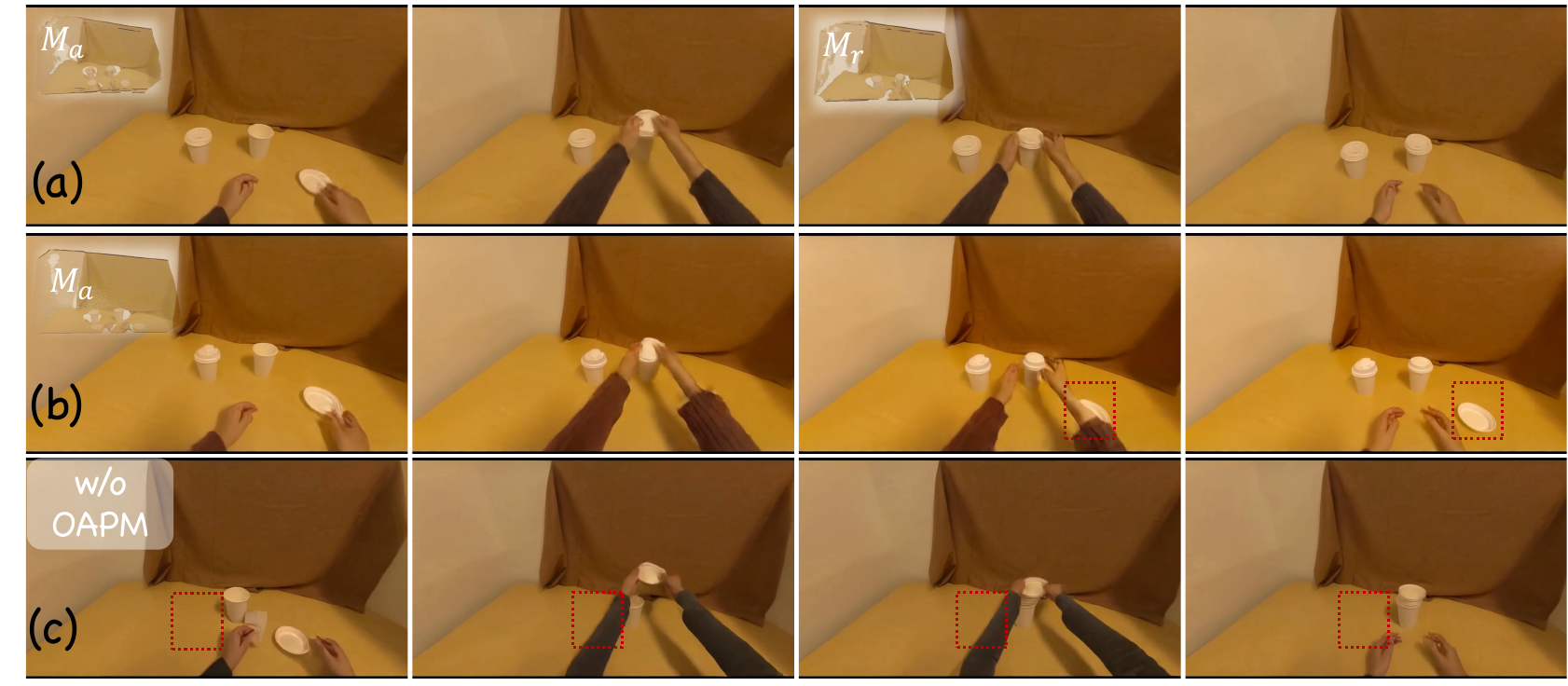}
    \caption{OAPM preserves persistent scene content and updates the current interaction state periodically.}
    \label{fig:oapm_qualitative_ablation}
\end{figure}

\paragraph{Qualitative scene-memory ablation.}
The anchor and recent slots in OAPM play complementary roles, as demonstrated in Figure~\ref{fig:oapm_qualitative_ablation}. The full model in (a) keeps the immutable $\mathcal{M}_a$ to preserve the clean scene layout and persistent object identity while replacing $\mathcal{M}_r$ from the most recent committed block to track the current interaction state. The anchor-only variant in (b) remains tied to stale object states, whereas removing OAPM in (c) allows objects to disappear or drift, as highlighted by the red boxes. Maintaining both slots therefore avoids overwriting the stable scene prior without sacrificing online state updates.

\subsection{Downstream WAM Generalization}

\begin{figure}[H]
    \centering
    \includegraphics[width=\columnwidth]{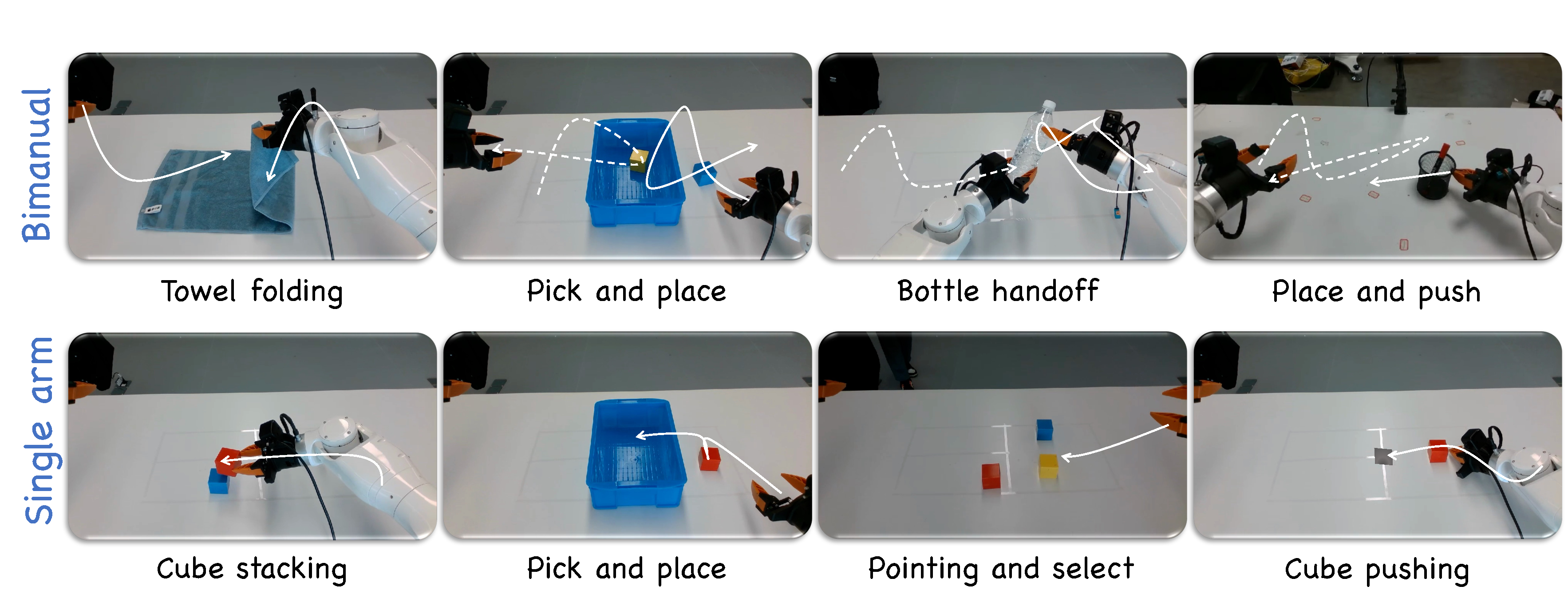}
    \caption{Eight real-robot tasks used to evaluate downstream WAM generalization.}
    \label{fig:downstream_tasks}
\end{figure}

Our eight-task real-robot suite spans precise contact, multi-stage execution, and coordination across embodiments, as summarized in Figure~\ref{fig:downstream_tasks}, where arrows indicate the intended motion. The bimanual tasks test coordinated folding, sequential pick-and-place, object handoff, and sequential place-and-push, while the single-arm tasks test stacking, pick-and-place, target selection, and pushing.

To test whether the generated rollouts provide useful supervision beyond perceptual gains, we fix the downstream WAM architecture and training schedule and vary only the training-data composition. Each real trajectory is converted into a first-frame scene anchor, camera trajectory, action skeleton or end-effector track, and language prompt; \method{} then resimulates it under edited appearance or scene conditions while preserving the action-label space. Augmenting 400 real trajectories with 400 synthetic trajectories raises OOD success from 77\% to 84\% on single-arm tasks and from 53\% to 70\% on dual-arm tasks, as shown in Table~\ref{tab:downstream_real}. It also reduces the corresponding ID-to-OOD losses from 7 to 4 points and from 19 to 6 points. Because the policy and optimization are held fixed, these gains demonstrate that \method{} supplies complementary variation that improves real-robot generalization under held-out appearances and layouts.

\begin{table}[t]
\centering
\scriptsize
\setlength{\tabcolsep}{2pt}
\renewcommand{\arraystretch}{1.08}
\resizebox{\columnwidth}{!}{%
\begin{tabular}{@{}llcccc@{}}
\toprule
Training data &
Split &
Single-arm SR$\uparrow$ &
SR Loss$\downarrow$ &
Dual-arm SR$\uparrow$ &
SR Loss$\downarrow$ \\
\midrule
\multirow{2}{*}{400 real} & ID & 84.0 & \multirow{2}{*}{\textcolor{gray}{$\downarrow$7.0}} & 72.0 & \multirow{2}{*}{\textcolor{gray}{$\downarrow$19.0}} \\
& OOD & 77.0 & & 53.0 & \\
\midrule
\multirow{2}{*}{400 synth.} & ID & 84.0 & \multirow{2}{*}{\textcolor{gray}{$\downarrow$8.0}} & 69.0 & \multirow{2}{*}{\textcolor{gray}{$\downarrow$13.0}} \\
& OOD & 76.0 & & 56.0 & \\
\midrule
\multirow{2}{*}{\makecell[l]{400 real +\\400 synth.}} & ID & \textbf{88.0} & \multirow{2}{*}{\textcolor{gray}{\textbf{$\downarrow$4.0}}} & \textbf{76.0} & \multirow{2}{*}{\textcolor{gray}{\textbf{$\downarrow$6.0}}} \\
& OOD & \textbf{84.0} & & \textbf{70.0} & \\
\bottomrule
\end{tabular}}
\caption{Real-robot task success rates (\%). SR denotes Success Rate; ID and OOD denote in-distribution and out-of-distribution evaluation, respectively.}
\label{tab:downstream_real}
\end{table}

\begin{figure}[!t]
    \centering
    \includegraphics[width=\columnwidth]{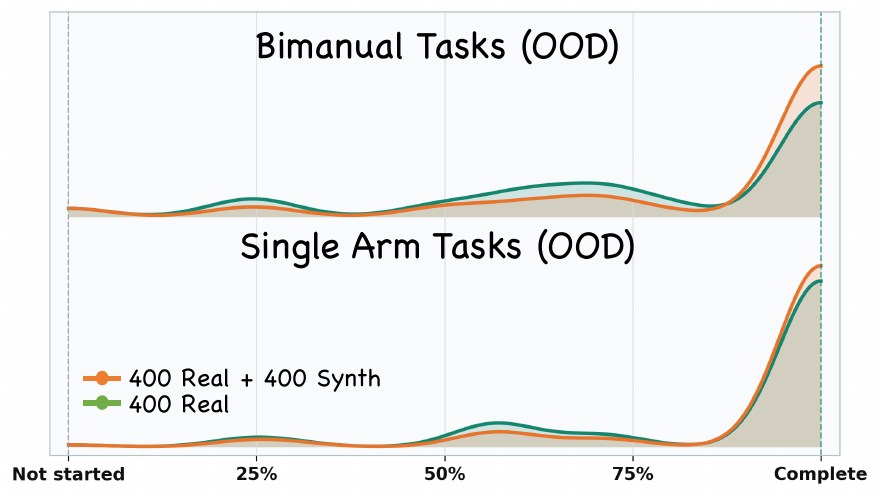}
    \caption{OOD task progress with and without \method-generated training data.}
    \label{fig:ood_stage_progress}
\end{figure}

\paragraph{OOD stage progress.}
Synthetic augmentation improves not only binary OOD success but also how far failed rollouts progress, as demonstrated in Figure~\ref{fig:ood_stage_progress}. The analysis resolves each rollout into its furthest stage over 25 trials per task; the rightmost mode is full task completion, and modes to its left denote approach, grasp, transport, and placement progress before failure. With 400 real trajectories alone, failures are distributed across the intermediate stages. Fine-tuning the downstream WAM with 400 additional \method-generated trajectories shifts both distributions toward later stages and complete executions: average completion rises from 53\% to 70\% for the bimanual suite and from 77\% to 84\% for the single-arm suite. This stage-wise shift independently corroborates the success-rate evidence in Table~\ref{tab:downstream_real}, showing that augmentation improves partial execution even when a rollout does not fully succeed.

\section{Conclusion}
\label{sec:conclusion}

We introduced \method{} as an autoregressive video generation model with OAPM and A3D-RoPE to address the scarcity of diverse egocentric manipulation data for embodied AI. OAPM combines a persistent 3D scene anchor with online recent-state refresh, while A3D-RoPE injects 3D geometry of the end-effector into gated cross-attention. Together, they produce high-quality egocentric rollouts with stronger visual fidelity, geometric stability, and action alignment. It allows for spanding current scarce video-action pairs with various attributes for egocentric data augmentation.  Beyond generation quality, current real data with our EgoGenesis-augmented data, can be used to train the downstream World Action Model to improve its generalization performance on both single- and dual-arm real-robot tasks. These results show that high-quality egocentric synthesis can expand limited real data and provide useful supervision for downstream WAM generalization.

{
    \small
    \bibliographystyle{ieeenat_fullname}
    \bibliography{aaai2027}
}

\clearpage
\appendix
\raggedbottom
\setcounter{topnumber}{4}
\setcounter{bottomnumber}{2}
\setcounter{totalnumber}{6}
\setcounter{dbltopnumber}{3}
\renewcommand{\topfraction}{0.95}
\renewcommand{\bottomfraction}{0.85}
\renewcommand{\textfraction}{0.05}
\renewcommand{\floatpagefraction}{0.75}
\renewcommand{\dbltopfraction}{0.95}
\renewcommand{\dblfloatpagefraction}{0.75}
\setlength{\textfloatsep}{8pt plus 2pt minus 2pt}
\setlength{\dbltextfloatsep}{8pt plus 2pt minus 2pt}
\makeatletter
\setlength{\@dblfptop}{0pt}
\setlength{\@dblfpbot}{0pt plus 1fil}
\makeatother
\twocolumn[
\centering
\Large
\textbf{Supplementary Material for\\
EgoGenesis: Egocentric World-Action Modeling with Online Anchored Projective Memory and Action-3D RoPE}\\
\vspace{0.5em}
\normalsize
\begin{tabular}{c}
Zexuan Yan, Yuzhou Wu, Yue Ma, Zonghang He, Kaibo Yin, Xiaobing Tu\\
Yinggui Wang, Jinkui Ren, Xiantao Zhang, Shijian Wang, Jinghong Liu, Linfeng Zhang$^{\dagger}$
\end{tabular}\par
\vspace{1.0em}
]

\addtocontents{toc}{\protect\setcounter{tocdepth}{2}}
\tableofcontents

\section{Technical Details for OAPM and A3D-RoPE}
\label{sec:supp-geometry}

\subsection{OAPM Encoding, Read, and Refresh}
OAPM treats the immutable anchor $\mathcal{M}_a$ and online recent memory $\mathcal{M}_r^b$ as abstract slots. Before block $b$, the pretrained VGGT-$\Omega$ encodes their concatenation, and its 3D scene reconstruction features are used directly as the scene embedding:
\begin{equation}
\mathbf{M}_b=\operatorname{SceneEncode}_{\Omega}
\!\left(\mathcal{M}_a\oplus\mathcal{M}_r^b\right).
\label{eq:supp-oapm-encode}
\end{equation}
The resulting tokens carry scene features, reference-frame 3D coordinates, and confidence. The memory read uses the same gated cross-attention form as the main paper:
\begin{equation}
\begin{aligned}
Q&=W_QH, & K&=W_K\mathbf{M}_b,\\
V&=W_V\mathbf{M}_b,\\
H&\leftarrow H+{}\\
&\quad\operatorname{GatedCrossAttn}(Q,K,V).
\end{aligned}
\label{eq:supp-oapm-read}
\end{equation}

In online mode, after every $s_r$ committed AR blocks, the pipeline decodes the causally visible latent prefix and uses its most recent RGB frame to construct a new snapshot:
\begin{equation}
\begin{aligned}
\mathcal{M}_{r}^{b+1}
&=E_{3D}\!\bigl(\operatorname{RecentFrame}\bigl(\\
&\qquad D_{\mathrm{vae}}(Z_{\le b})\bigr)\bigr),\\
b&\equiv0\pmod {s_r}.
\end{aligned}
\label{eq:supp-oapm-online}
\end{equation}
The anchor is unchanged and the recent snapshot is replaced, combining stable reference geometry with the latest scene state.

\subsection{A3D-RoPE Encoding and Coordinate Construction}
The rendered-skeleton VAE latent is patchified on the video grid. Let $\mathcal{I}_b$ denote the patches covered by the skeleton in block $b$; $\mathbf{X}_b$ collects their anchor-frame metric 3D coordinates, and $X_a$ is one component along axis $a\in\{x,y,z\}$. Queries originate from the video hidden states, whereas keys and values originate from the patch-aligned skeleton tokens. In the following, $Q$, $K$, and $V$ refer only to the QKV entries indexed by $\mathcal{I}_b$ after their standard linear projections, rather than to QKV over the full patch grid. A3D-RoPE splits the selected query and key channels into three spatial groups,
\begin{equation}
Q=[Q^x\,\|\,Q^y\,\|\,Q^z],
\qquad
K=[K^x\,\|\,K^y\,\|\,K^z].
\label{eq:supp-a3d-split}
\end{equation}
Within each group, adjacent channels form standard two-dimensional RoPE pairs. If $M_a$ pairs are assigned to axis $a\in\{x,y,z\}$, pair $m$ uses
\begin{equation}
\begin{aligned}
\theta_{a,m}(X_a)&=sX_a\kappa^{-m/M_a},\\
\begin{bmatrix}u'\\v'\end{bmatrix}
&=
\begin{bmatrix}
\cos\theta_{a,m}(X_a)&-\sin\theta_{a,m}(X_a)\\
\sin\theta_{a,m}(X_a)&\cos\theta_{a,m}(X_a)
\end{bmatrix}
\begin{bmatrix}u\\v\end{bmatrix},
\end{aligned}
\label{eq:supp-a3d-rotate}
\end{equation}
Here, $\theta_{a,m}(X_a)$ is the rotation angle induced by $X_a$ at the $m$-th RoPE frequency, with $s=4$ and $\kappa=10^4$. For each supported patch in block $b$, its entry in $\mathbf{X}_b$ is represented by $(X_x,X_y,X_z)$ in the anchor frame. Therefore, $X_a$ is one axis component of $\mathbf{X}_b$, and applying the axis-wise rotation to every supported patch yields $R_{\mathbf{X}_b}$. The video and action features are then rotated before gated cross-attention:
\begin{equation}
\begin{aligned}
\widetilde Q&=R_{\mathbf{X}_b}(Q),
&\widetilde K&=R_{\mathbf{X}_b}(K),\\
H_{\mathcal{I}_b}&\leftarrow H_{\mathcal{I}_b}+\operatorname{GatedCrossAttn}(\widetilde Q,\widetilde K,V).
\end{aligned}
\label{eq:supp-a3d-attn}
\end{equation}
The update is written back only at $\mathcal{I}_b$ because these are the patches with valid action coordinates; unselected background states remain unchanged.

We next detail how the skeleton-supported coordinates $\mathbf{X}_{tp}$ are constructed. For skeleton edge $e=(j,k)$, let $\mathbf{u}_{tj}$ and $\mathbf{u}_{tk}$ be normalized image coordinates and $d_{tj},d_{tk}$ their valid depths. The projection of patch center $\mathbf{u}_p$ onto the edge is
\begin{equation}
\alpha_{tpe}=\operatorname{clip}_{[0,1]}
\frac{(\mathbf{u}_p-\mathbf{u}_{tj})^{\top}
(\mathbf{u}_{tk}-\mathbf{u}_{tj})}
{\|\mathbf{u}_{tk}-\mathbf{u}_{tj}\|_2^2},
\label{eq:supp-a3d-alpha}
\end{equation}
where the numerator denotes the inner product. The perspective-correct edge depth and Gaussian support are
\begin{equation}
\begin{aligned}
d_{tpe}&=\frac{d_{tj}d_{tk}}
 {(1-\alpha_{tpe})d_{tk}+\alpha_{tpe}d_{tj}},\\
w_{tpe}&=\mathbf{1}_{e}\exp\!\left(
 -\frac{\|\mathbf{u}_p-[(1-\alpha_{tpe})\mathbf{u}_{tj}
 +\alpha_{tpe}\mathbf{u}_{tk}]\|_2^2}{2r_{te}^2}\right),\\
d_{tp}&=\frac{\sum_e w_{tpe}d_{tpe}}{\sum_e w_{tpe}}.
\end{aligned}
\label{eq:supp-a3d-raster}
\end{equation}
Here $\mathbf{1}_e$ requires valid endpoints, and
$r_{te}=r_0+0.2\|\mathbf{u}_{tk}-\mathbf{u}_{tj}\|_2$ is the adaptive tube radius (with $r_0=0.10$). The supported patch ray is unprojected and expressed in the reference-camera frame as
\begin{equation}
\mathbf{X}_{tp}=\left[
\mathbf{V}_{\mathrm{ref}}\mathbf{V}_t^{-1}
\begin{bmatrix}d_{tp}\mathbf{K}_t^{-1}\bar{\mathbf{u}}_p\\1\end{bmatrix}
\right]_{1:3}.
\label{eq:supp-a3d-unproject}
\end{equation}
Patches without valid skeleton support are excluded from A3D-RoPE cross-attention.

\section{Training Corpus and Autoregressive Procedure}

\subsection{Training Data}
We train on a source-balanced 210K-clip egocentric corpus. EgoDex and AgiBot contribute 100K clips each, complemented by 4K RoboTwin clips, 5K Real-world Ego clips, and 1K DexJoCo clips. Real-world Ego combines teleoperated rollouts in the downstream task environments, egocentric human-hand interactions, and first-person recordings collected on different robot embodiments. Every source is converted to a unified camera-and-pose conditioning interface. All training and test splits are disjoint at the clip and trajectory levels, and videos are standardized to 81 frames at 16 FPS and $832\times480$. All \method{} training runs use NVIDIA A100 GPUs.

\begin{figure}[H]
    \centering
    \includegraphics[width=\columnwidth]{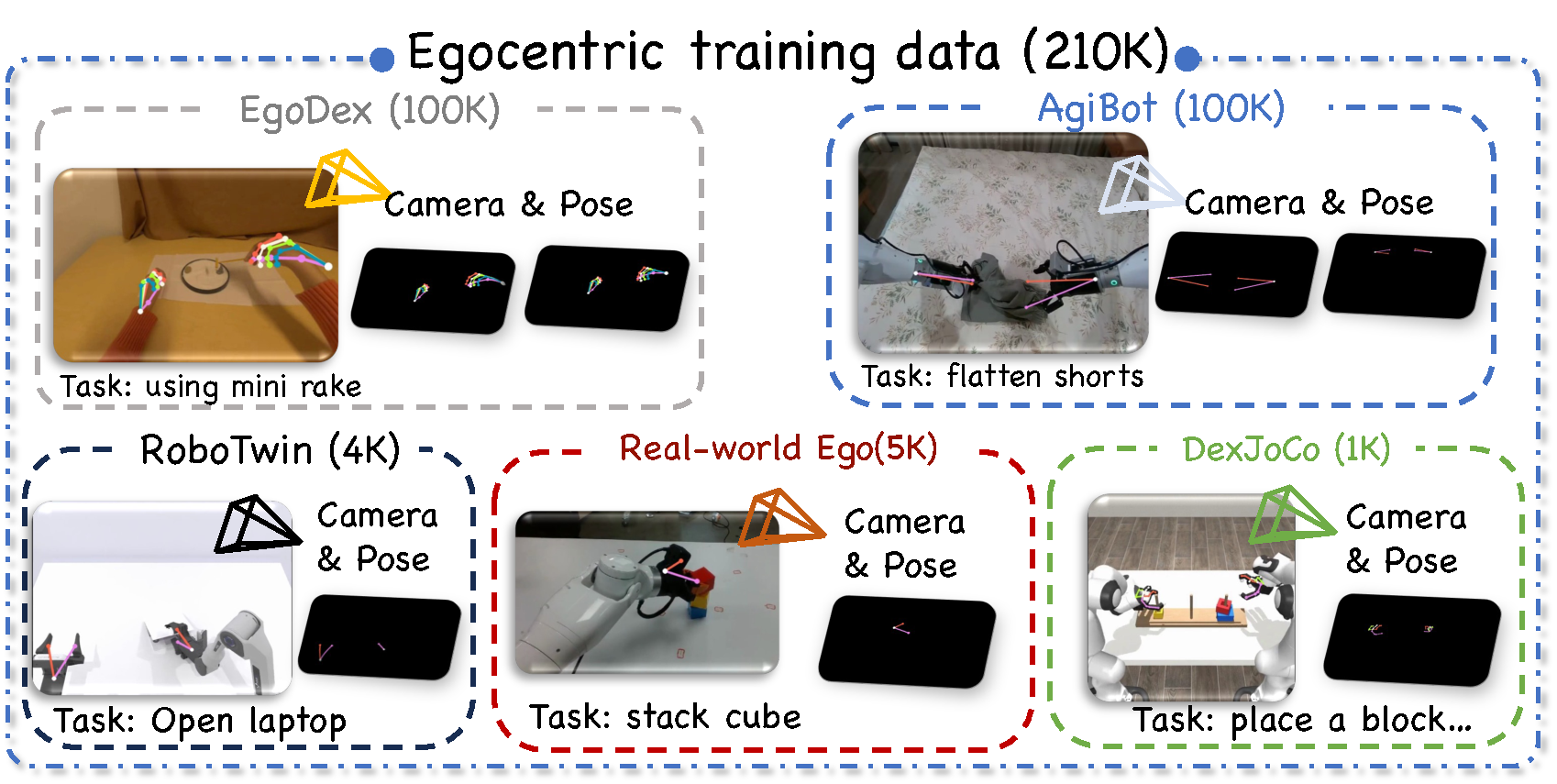}
    \caption{Composition of the 210K-clip egocentric training corpus. Every source is converted to a unified camera-and-pose conditioning interface.}
    \label{fig:supp_training_data}
\end{figure}

\subsection{Autoregressive Training and Inference}
Starting from the pretrained video prior, we train \method{} with block-causal attention: two context blocks are teacher-forced and one target block is denoised at a time, with random target windows and online OAPM refresh. Training uses the source-balanced mixture above.

For AR training, the clean context blocks are written into $\mathcal{K}_{<b}$ and the target block is optimized with the flow-matching objective
\begin{equation}
\mathcal{L}_{\mathrm{AR}}=\mathbb{E}_{b,t}\left[\left\|\widehat v_b^t-(\varepsilon_b-Z_b)\right\|_2^2\right],
\label{eq:ar-training}
\end{equation}
which exposes the model to the same causal context used at inference. At inference, each chunk is sampled from noise, committed to the cache, and used to condition the next chunk:
\begin{equation}
\widehat Z_b=\operatorname{FlowSample}(\mathbf{c}_b,\mathcal{K}_{<b}),
\qquad
\mathcal{K}_{<b+1}=\mathcal{K}_{<b}\mathbin{\oplus}\widehat Z_b.
\label{eq:ar-inference}
\end{equation}
The initial frame initializes $\mathcal{M}_{a}$, the recent memory $\mathcal{M}_{r}$ starts empty, and Eq.~\eqref{eq:supp-oapm-online} refreshes it at the configured stride. The final rollout is decoded after all chunks are committed.

\subsection{AR Generation with OAPM and A3D-RoPE}
The complete blockwise generation procedure is given in Algorithm~\ref{alg:egogenesis-ar}.

\begingroup
\renewcommand{\algorithmicdo}{}
\renewcommand{\algorithmicthen}{}
\newcommand{\righttricomment}[1]{\item[]\hfill$\triangleright$~#1}
\begin{algorithm}[H]
    \caption{AR Generation with OAPM and A3D-RoPE}
    \label{alg:egogenesis-ar}
    \small
    \begin{algorithmic}[1]
    \Require Initial frame $I_0$, prompt $y$, block conditions $\{\mathcal{C}_b,S_b,\mathbf{X}_b\}_{b=1}^{B}$, refresh stride $s_r$.
    \Ensure Generated rollout $\widehat V$.

    \State $\mathcal{M}_{a}\gets E_{3D}(I_0)$; $\mathcal{M}_{r}^{1}\gets\varnothing$; $\mathcal{K}_{<1}\gets\varnothing$
    \righttricomment{\textbf{Stage 1:} \textit{Condition initialization}}
    \For{autoregressive block $b=1,\ldots,B$}
        \State $Z_b^{1}\sim\mathcal{N}(0,\mathbf{I})$
        \State $\mathbf{M}_b\gets\operatorname{SceneEncode}_{\Omega}(\mathcal{M}_{a}\oplus\mathcal{M}_{r}^{b})$
        \State $\mathcal{I}_b\gets\operatorname{SkeletonSupport}(S_b)$
        \righttricomment{\textbf{Stage 2:} \textit{Blockwise flow integration}}
        \For{flow time $t$ from $1$ to $0$}
            \State $H\gets\operatorname{PatchEmbed}(Z_b^{t})+\operatorname{TimeEmbed}(t)$
            \For{each DiT layer}
                \State $H\gets\operatorname{DiTLayer}(H;y,\mathcal{C}_b,\mathcal{K}_{<b})$
                \If{the layer contains an OAPM adapter}
                    \State $H\gets H+\operatorname{GatedCrossAttn}_{\mathrm{OAPM}}(H,\mathbf{M}_b)$
                \EndIf
                \If{the layer contains an A3D-RoPE adapter}
                    \State $H_{\mathcal{I}_b}\gets H_{\mathcal{I}_b}+\operatorname{GatedCrossAttn}_{\mathrm{A3D}}(H_{\mathcal{I}_b},S_{b,\mathcal{I}_b},\mathbf{X}_b)$
                \EndIf
            \EndFor
            \State $\widehat v_b^t\gets\operatorname{Head}(H)$; $Z_b^{t-\Delta t}\gets Z_b^t-\Delta t\,\widehat v_b^t$
        \EndFor
        \State $Z_b\gets Z_b^{0}$; append $Z_b$ to $\mathcal{K}_{<b+1}$
        \righttricomment{\textbf{Stage 3:} \textit{Causal commit and memory refresh}}
        \State $\mathcal{M}_{r}^{b+1}\gets\mathcal{M}_{r}^{b}$
        \If{$b\bmod s_r=0$}
            \State $\widehat I_b^{\mathrm{rec}}\gets\operatorname{RecentFrame}(D_{\mathrm{vae}}(Z_{\le b}))$
            \State $\mathcal{M}_{r}^{b+1}\gets E_{3D}(\widehat I_b^{\mathrm{rec}})$
        \EndIf
    \EndFor
    \State \Return $\widehat V\gets D_{\mathrm{vae}}([Z_1,\ldots,Z_B])$
    \end{algorithmic}
\end{algorithm}
\endgroup

\section{Complete Component Ablation}
The full ablation reports action alignment, physical faithfulness, and temporal consistency in addition to the compact visual-fidelity columns retained in the main paper.
\begin{table*}[t]
\centering
\scriptsize
\setlength{\tabcolsep}{4.2pt}
\renewcommand{\arraystretch}{1.08}
\resizebox{\textwidth}{!}{%
\begin{tabular}{>{\raggedright\arraybackslash}p{3.8cm}ccccccc}
\toprule
Component Setting &
\makecell{PSNR$\uparrow$} &
\makecell{SSIM$\uparrow$} &
\makecell{LPIPS$\downarrow$} &
\makecell{Kpt.Err$\downarrow$} &
\makecell{Phys.Faith$\uparrow$} &
\makecell{Subj. Cons.$\uparrow$} &
\makecell{Bg. Cons.$\uparrow$} \\
\midrule
\textbf{Wan2.2-5B-Control-AR} & 19.9238 & 0.7812 & 0.3028 & 0.07723 & 0.7796 & 0.8337 & 0.9316 \\
\midrule
\multicolumn{8}{l}{\textit{Scene memory} (A3D-RoPE fixed)} \\
First-frame anchor only & 20.4135 & 0.8385 & 0.2533 & 0.0532 & 0.8037 & 0.8847 & 0.9532 \\
\textbf{$+$ Recent refresh} & \textbf{21.8609} & \textbf{0.8509} & \textbf{0.2399} & \textbf{0.0501} & 0.8278 & 0.8923 & 0.9546 \\
\midrule
\multicolumn{8}{l}{\textit{Positional encoding} (OAPM fixed)} \\
RoPE & 21.4250 & 0.8198 & 0.2838 & 0.07719 & 0.8182 & 0.8837 & 0.9353 \\
PRoPE & 21.8421 & 0.8408 & 0.2481 & 0.06135 & 0.8255 & 0.8919 & 0.9481 \\
\textbf{A3D-RoPE} & \textbf{21.8609} & \textbf{0.8509} & \textbf{0.2399} & \textbf{0.0501} & 0.8278 & 0.8923 & 0.9546 \\
\bottomrule
\end{tabular}}
\caption{Core-component ablations on Wan2.2-5B-Control. Within each block, the complementary component is fixed to its full configuration.}
\label{tab:ablation}
\end{table*}

\section{Detailed Downstream Real-Robot Results}

\begin{figure*}[t]
    \centering
    \includegraphics[width=\textwidth]{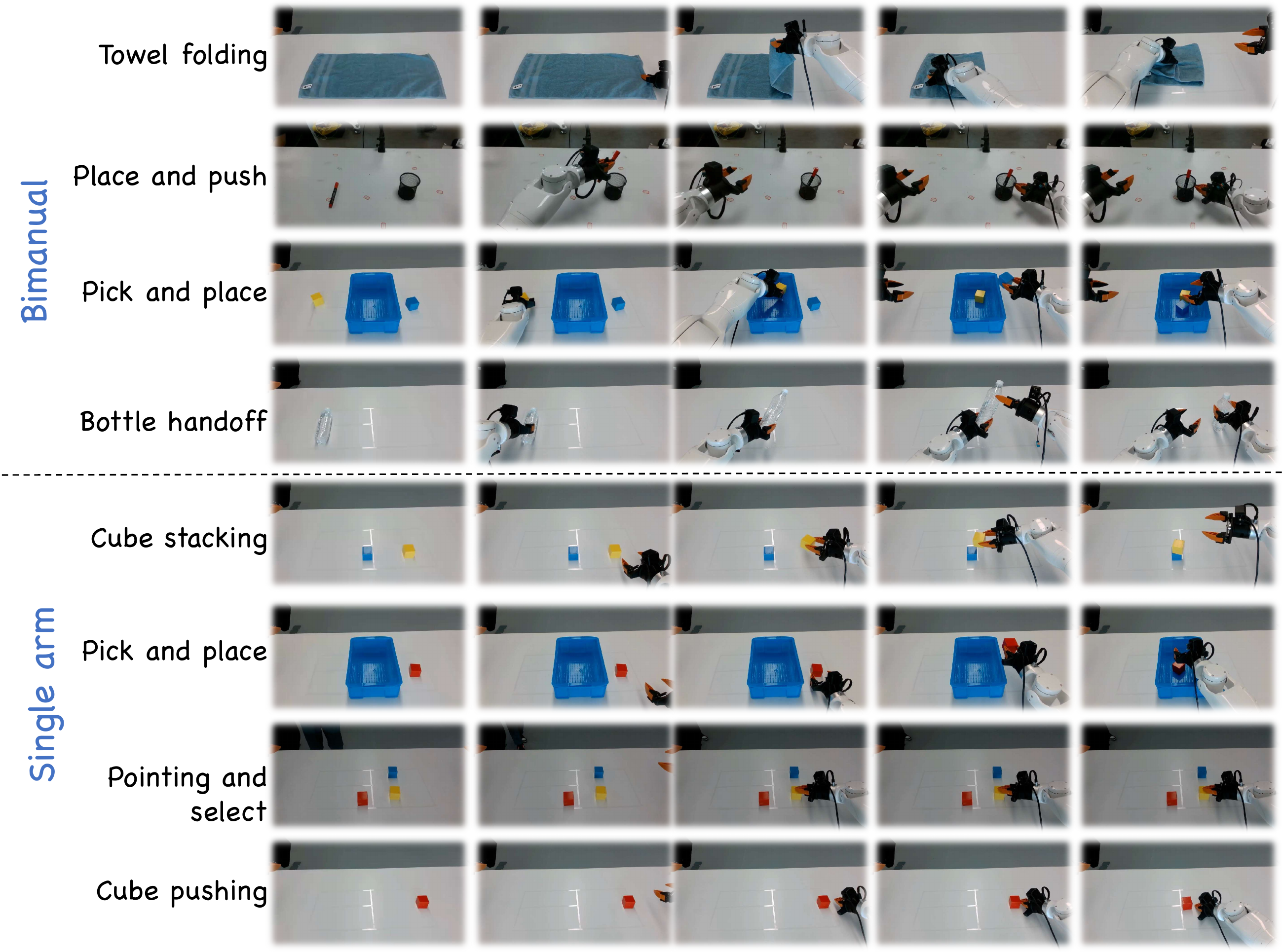}
    \caption{Detailed execution sequences for four bimanual and four single-arm real-robot tasks. Five checkpoints per row show the progression from the initial configuration to the task-specific success state.}
    \label{fig:supp_task_execution}
\end{figure*}

The complete execution sequences in Figure~\ref{fig:supp_task_execution} provide a task-level view of the real-robot evaluation summarized in main-paper Table~3. Each row contains five checkpoints spanning initialization, approach, contact, intermediate transition, and the terminal state. Bimanual success requires a stable towel fold, placing the pen-like object before pushing its receiving cup, placing both objects into the bin, or completing the bottle handoff. Single-arm success requires a stable two-cube stack, placing the cube inside the bin, selecting the instructed target without disturbing distractors, or pushing the cube fully into the marked region.

Main-paper Table~3 reports aggregate success rates for the single-arm and dual-arm suites, while Supplementary Table~\ref{tab:supp_downstream_breakdown} provides the corresponding per-task results under the same policy architecture and training schedule. We compare 400 real trajectories, 400 synthetic trajectories, and their combination. ID trials retain training-time object appearances and layouts, whereas OOD trials hold out object appearances and initial or goal layouts. Each task is evaluated over 25 trials, so its success rate changes in 4-point increments; each aggregate is the exact mean over four tasks, or equivalently the success rate over 100 trials.

\subsection{Tianji M6 Platform, Tasks, and Evaluation Protocol}

\begin{figure}[!t]
    \centering
    \includegraphics[width=\columnwidth]{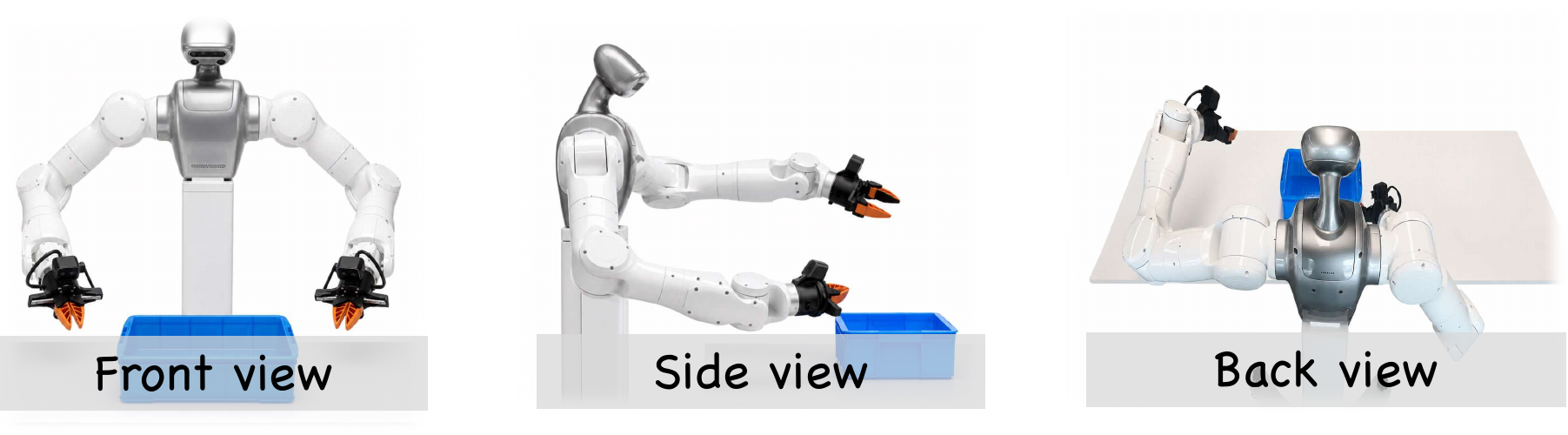}
    \caption{Tianji M6 real-robot environment from front, side, and back views. The two gripper-equipped arms share a tabletop workspace observed by the head-mounted egocentric camera.}
    \label{fig:supp_robot_environment}
\end{figure}

All real-robot trials use the Tianji M6 platform shown in Figure~\ref{fig:supp_robot_environment}. Its two gripper-equipped arms operate over an overlapping tabletop workspace, while the head-mounted camera provides the egocentric observation used by the policy. Single-arm trials activate one arm and keep the other outside the task workspace; bimanual trials coordinate both arms. The robot, camera mounting, and workspace remain fixed across training-data settings and ID/OOD evaluation, so the comparison isolates changes in object appearance, initial configuration, and target layout.

The single-arm suite includes Cube Stacking into a stable tower, Pick and Place into a bin, Pointing and Select without disturbing distractors, and Cube Pushing into a taped region. The dual-arm suite includes Towel Folding with one arm anchoring the cloth, Pick and Place of two objects into a central bin, Bottle Handoff followed by upright placement, and Place and Push of a pen-like object and its receiving cup.

\subsubsection{Task Stages and Completion Criteria}

Each task is divided into ordered interaction stages to make the execution progress and terminal success criterion explicit. The percentages below indicate normalized trajectory progress rather than success rates; \textit{Done} denotes the task-specific terminal state at 100\%.

\begin{table*}[!t]
\centering
\footnotesize
\renewcommand{\arraystretch}{0.98}
\begin{tabular}{@{}>{\raggedright\arraybackslash}p{0.18\linewidth}>{\raggedright\arraybackslash}p{0.18\linewidth}>{\centering\arraybackslash}p{0.10\linewidth}>{\raggedright\arraybackslash}p{0.46\linewidth}@{}}
\toprule
\textbf{Task} & \textbf{Stage} & \textbf{Progress} & \textbf{Completion criterion} \\
\midrule

\textbf{Towel Folding} & S1: Approach and grasp & 25\% & Both grippers approach the towel, stabilize one side, and grasp the side to be folded. \\*
& S2: Lift and fold & 62\% & The right side of the towel is lifted and folded toward the left. \\*
& \textit{Done}: Align and flatten & 100\% & The folded edge is adjusted and flattened to form a stable, aligned fold. \\
\midrule

\textbf{Bimanual Pick and Place} & S1: Grasp both cubes & 28\% & The left and right grippers approach and grasp the cubes on their respective sides. \\*
& S2: Place the blue cube & 58\% & The blue cube is transported into the central bin. \\*
& S3: Transport the red cube & 80\% & The red cube is lifted and moved above the bin. \\*
& \textit{Done}: Place the red cube & 100\% & The red cube is placed in the bin so that both cubes reach their target location. \\
\midrule

\textbf{Bottle Handoff} & S1: Grasp the bottle & 22\% & The delivering gripper grasps the horizontally placed bottle. \\*
& S2: Lift and rotate & 48\% & The bottle is lifted from the table and rotated toward an upright pose. \\*
& S3: Transfer control & 70\% & The receiving gripper secures the bottle and completes the handoff. \\*
& \textit{Done}: Place upright & 100\% & The receiving gripper places the bottle upright at the target location. \\
\midrule

\textbf{Place and Push} & S1: Grasp the pen and cup & 23\% & The two grippers take control of the pen and cup, respectively. \\*
& S2: Lift and align & 52\% & The pen is moved above the cup opening while the other gripper stabilizes the cup. \\*
& S3: Insert the pen & 72\% & The pen is inserted into the cup and reaches a stable state. \\*
& \textit{Done}: Push the cup & 100\% & The cup containing the pen is pushed to the target location. \\
\midrule
\textbf{Cube Stacking} & S1: Approach and grasp & 24\% & The gripper approaches and grasps the yellow cube. \\*
& S2: Lift and transport & 56\% & The yellow cube is moved above the blue cube. \\*
& \textit{Done}: Align and stack & 100\% & The cubes are aligned to form a stable two-level stack. \\
\midrule

\textbf{Single-Arm Pick and Place} & S1: Approach and grasp & 26\% & The gripper approaches and grasps the cube outside the bin. \\*
& S2: Lift and transport & 58\% & The cube is moved above the bin and aligned with its opening. \\*
& \textit{Done}: Place in the bin & 100\% & The cube is released into the bin and reaches the target state. \\
\midrule

\textbf{Point and Select} & S1: Approach the target & 30\% & The gripper moves toward the specified red target without contacting the other cubes. \\*
& S2: Align the pointing pose & 68\% & The pointing direction and position are aligned above the target. \\*
& \textit{Done}: Point and hold & 100\% & A clear and stable pointing pose is maintained near the red target. \\
\midrule

\textbf{Cube Pushing} & S1: Approach and contact & 30\% & The gripper approaches the red cube and establishes pushing contact. \\*
& S2: Push toward the target & 72\% & Continuous contact moves the cube toward the black marked region. \\*
& \textit{Done}: Stabilize in the target & 100\% & The cube lies fully within the target region and remains stable. \\
\bottomrule
\end{tabular}
\caption{Stage-wise execution protocol for the eight downstream real-robot tasks.}
\label{tab:supp_task_stages}
\end{table*}

\begin{table*}[!htbp]
\centering
\small
\setlength{\tabcolsep}{7pt}
\renewcommand{\arraystretch}{1.03}
\textbf{(a) Single-arm tasks on Tianji M6}\par\vspace{2pt}
\resizebox{\textwidth}{!}{%
\begin{tabular}{lccccc}
\toprule
Training data &
\makecell{Cube Stacking} &
\makecell{Pick\&Place} &
\makecell{Pointing\\\& Select} &
\makecell{Cube Pushing} &
\makecell{Avg. SR$\uparrow$} \\
\midrule
\multicolumn{6}{l}{\textit{In-distribution (ID)}} \\
400 real & 76 & 84 & 92 & 84 & 84.0 \\
400 real + 400 synth. & \textbf{80} & \textbf{88} & \textbf{96} & \textbf{88} & \textbf{88.0} \\
400 synth. & 72 & 84 & \textbf{96} & 84 & 84.0 \\
\midrule
\multicolumn{6}{l}{\textit{Out-of-distribution (OOD)}} \\
400 real & 56 & 80 & 92 & 80 & 77.0 \\
400 real + 400 synth. & \textbf{72} & \textbf{84} & \textbf{96} & \textbf{84} & \textbf{84.0} \\
400 synth. & 52 & 80 & 92 & 80 & 76.0 \\
\bottomrule
\end{tabular}}

\vspace{7pt}
\textbf{(b) Dual-arm tasks on Tianji M6}\par\vspace{2pt}
\resizebox{\textwidth}{!}{%
\begin{tabular}{lccccc}
\toprule
Training data &
\makecell{Towel Folding} &
\makecell{Pick\&Place} &
\makecell{Bottle\\Handoff} &
\makecell{Place \&\\Push} &
\makecell{Avg. SR$\uparrow$} \\
\midrule
\multicolumn{6}{l}{\textit{In-distribution (ID)}} \\
400 real & 72 & 76 & 68 & 72 & 72.0 \\
400 real + 400 synth. & \textbf{76} & \textbf{80} & \textbf{72} & \textbf{76} & \textbf{76.0} \\
400 synth. & 68 & 72 & 68 & 68 & 69.0 \\
\midrule
\multicolumn{6}{l}{\textit{Out-of-distribution (OOD)}} \\
400 real & 52 & 56 & 52 & 52 & 53.0 \\
400 real + 400 synth. & \textbf{68} & \textbf{80} & \textbf{68} & \textbf{64} & \textbf{70.0} \\
400 synth. & 52 & 64 & 56 & 52 & 56.0 \\
\bottomrule
\end{tabular}}
\caption{Per-task real-robot success rates (\%), with 25 trials per task. Panels (a) and (b) exactly decompose the Single-arm SR and Dual-arm SR columns in main-paper Table~3; bold denotes the best training budget within each split.}
\label{tab:supp_downstream_breakdown}
\end{table*}

\subsection{Downstream WAM Training and Inference Details}
\label{sec:supp_downstream_training}

For each training-data setting, we independently initialize LingBot-VA from its official released checkpoint and fine-tune it on the corresponding real, synthetic, or mixed trajectory set. The downstream WAM therefore does not inherit weights from our trained Wan2.2-5B-Control generator. We keep the model architecture, optimization configuration, and inference settings identical across data settings, changing only the composition of the fine-tuning trajectories. Each sample contains three synchronized RGB observations from the head-mounted camera, the left-wrist camera, and the right-wrist camera. All observations are resized to $256\times256$. The temporal attention window is set to 30, and the video frame chunk size is set to 2.

The model maintains a canonical 30-dimensional action space. For the bimanual joint-control experiments, we select 16 active channels corresponding to seven left-arm joint dimensions, the left gripper, seven right-arm joint dimensions, and the right gripper. Using zero-based indexing, the selected channels are ordered as
\begin{equation}
[14{:}20,\;28,\;21{:}27,\;29].
\label{eq:supp_active_action_channels}
\end{equation}
The end-effector channels are therefore not used in these experiments. The action-per-frame factor is set to 16. Each active action dimension is normalized using its 1st and 99th percentiles computed from the corresponding training data.

We optimize the model using AdamW with a learning rate of $1\times10^{-5}$, coefficients $(\beta_1,\beta_2)=(0.9,0.95)$, and weight decay of $0.1$. The first 200 training steps are used for learning-rate warmup. The per-GPU batch size is 1, and gradients are accumulated over four iterations. LingBot-VA training and testing, including downstream real-robot inference, use NVIDIA H100 GPUs; training is distributed over eight H100 GPUs, resulting in an effective global batch size of 32. Each model is trained for 20,000 steps. During training, the language-condition embedding is replaced by the empty-text embedding with probability 0.1 for classifier-free guidance training. Checkpoints are saved every 1,000 steps. A complete training run takes approximately 96 hours on eight H100 GPUs.

During inference, the classifier-free guidance scales are set to 5 for the video branch and 1 for the action branch. We use 5 denoising steps for video generation and 10 denoising steps for action generation. The video and action SNR-shift parameters are set to 5.0 and 1.0, respectively. The video denoising process is not truncated. Table~\ref{tab:supp_downstream_hyperparameters} summarizes the complete configuration.

\begin{table*}[!t]
\centering
\caption{Training and inference configuration of the downstream WAM. The same configuration is used for the real, synthetic, and mixed training-data settings.}
\label{tab:supp_downstream_hyperparameters}
\small
\begin{tabular}{p{0.44\linewidth}p{0.44\linewidth}}
\toprule
\textbf{Hyperparameter} & \textbf{Setting} \\
\midrule
Pretrained initialization & Official LingBot-VA release checkpoint; independent initialization for each data setting \\
Observation cameras & Head, left wrist, and right wrist \\
Input resolution & $256\times256$ \\
Temporal attention window & 30 \\
Video frame chunk size & 2 \\
Canonical action dimension & 30 \\
Active action dimension & 16 \\
Active action channels & $[14{:}20,\,28,\,21{:}27,\,29]$ \\
Action-per-frame factor & 16 \\
Action normalization & Per-dimension 1st/99th percentiles \\
\midrule
Optimizer & AdamW \\
Learning rate & $1\times10^{-5}$ \\
Adam coefficients & $(0.9,0.95)$ \\
Weight decay & $0.1$ \\
Warmup steps & 200 \\
Per-GPU batch size & 1 \\
Gradient accumulation steps & 4 \\
Training hardware & 8 NVIDIA H100 GPUs \\
Effective global batch size & 32 \\
Training steps & 20,000 \\
Text-condition dropout probability & 0.1 \\
Checkpoint interval & 1,000 steps \\
Training time & Approximately 96 hours \\
\midrule
Video CFG scale & 5 \\
Action CFG scale & 1 \\
Video denoising steps & 5 \\
Action denoising steps & 10 \\
Video SNR shift & 5.0 \\
Action SNR shift & 1.0 \\
Video denoising truncation & None \\
\bottomrule
\end{tabular}
\end{table*}

\begin{figure*}[!t]
    \centering
    \includegraphics[width=\textwidth]{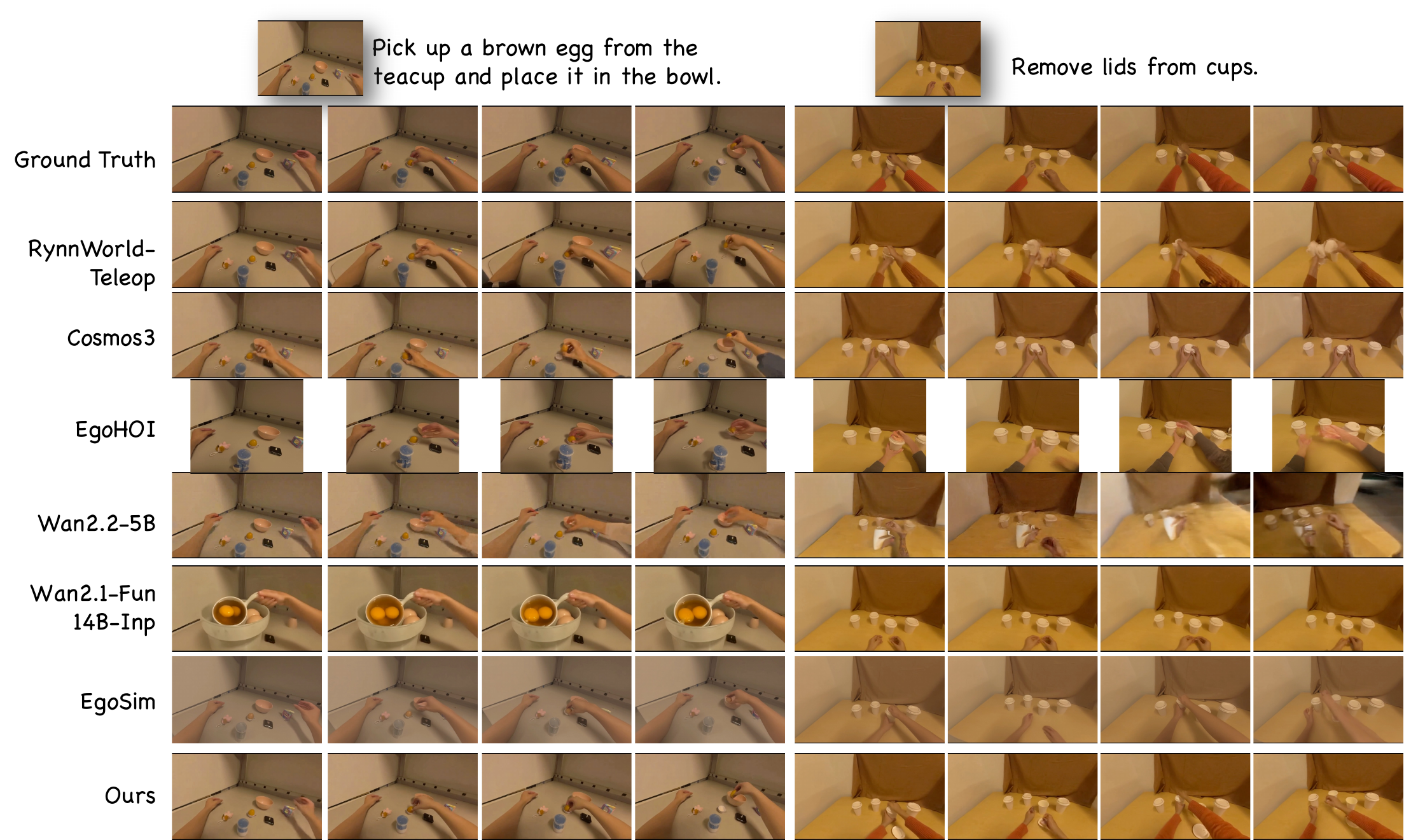}
    \caption{Additional comparisons on egg transfer and cup-lid removal; \method{} better preserves scene and hand identity while following the commanded contact sequence.}
    \label{fig:supp_qualitative_comparison}
\end{figure*}

\section{Additional Qualitative Comparisons}

Figure~\ref{fig:supp_qualitative_comparison} extends the main-paper comparison to egg transfer and cup-lid removal. In the egg task, \method{} preserves the tabletop objects and follows the hand from the teacup toward the bowl, whereas several baselines alter the setup, drift in hand/object geometry, or under-execute the transfer. In the lid-removal task, \method{} retains the cup arrangement and red-sleeved hand identity while producing successive contact-driven changes; competing generations more often merge hands with lids, deform cups, or show limited task progression.

\subsection{Cross-Embodiment Simulation in an Unseen Environment}

\begin{figure}[H]
    \centering
    \includegraphics[width=\columnwidth]{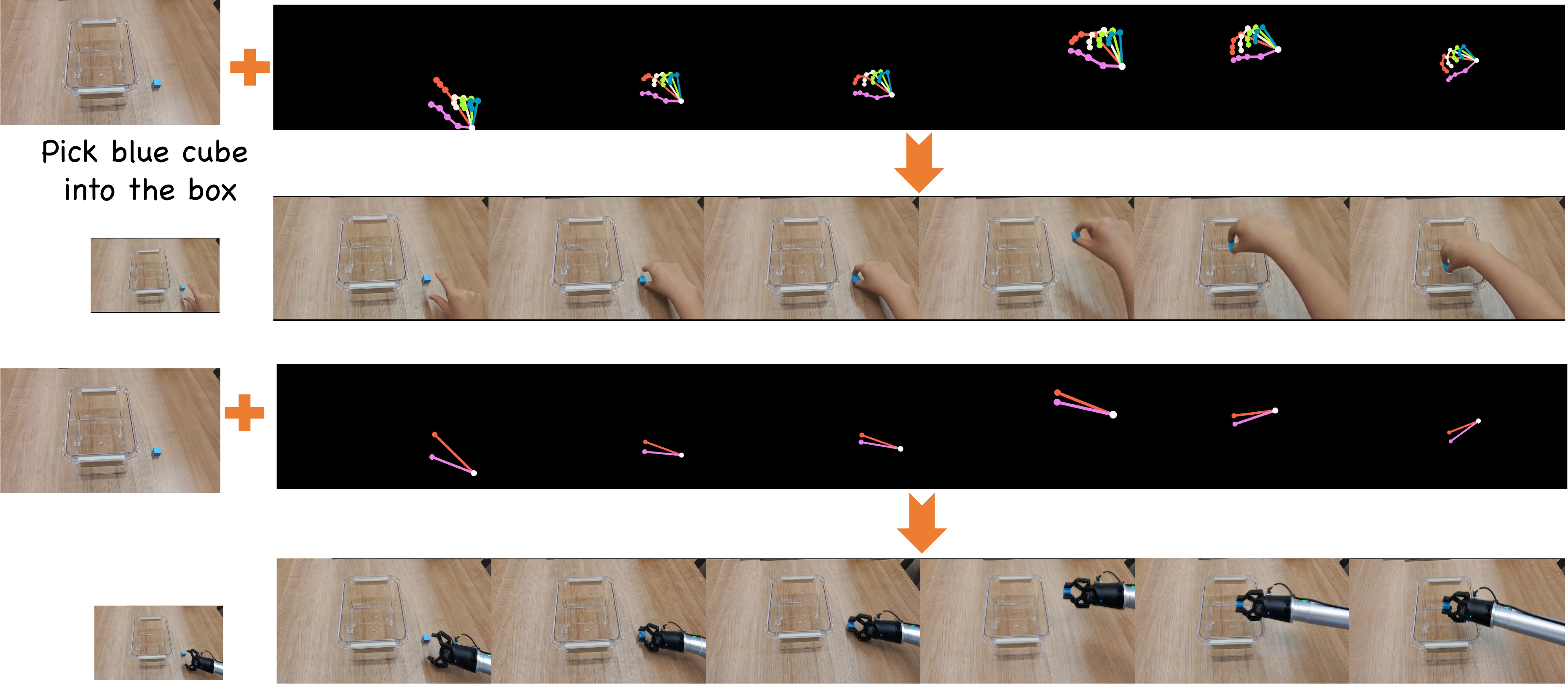}
    \caption{Cross-embodiment simulation in an unseen environment. Given the same initial scene and instruction, \method{} follows a full hand skeleton to generate human-hand manipulation (top) and a compact gripper skeleton to generate robot-gripper manipulation (bottom). The gripper skeleton is extracted from the index-finger and thumb trajectories of the hand skeleton.}
    \label{fig:supp_cross_embodiment}
\end{figure}

Figure~\ref{fig:supp_cross_embodiment} shows that \method{} can simulate manipulation in an environment not observed during training while preserving the specified action. Starting from the same unseen scene and instruction, the model generates both human-hand and robot-gripper executions. The gripper control skeleton in the lower example is obtained by extracting the index-finger and thumb trajectories from the full hand skeleton in the upper example, providing a compact action condition for cross-embodiment simulation.

\section{Evaluation Metrics and Kimi K2.7 Prompt}

\subsection{Main-Table Metrics}

All methods are evaluated on temporally aligned generated and reference clips under the same scene and action conditions. Frame-level values are first averaged within each sample; table entries are arithmetic means over non-empty per-sample values. The seven metrics reported in the main results and the supplementary component ablation are defined as follows.

\begin{table*}[!t]
\centering
\begin{tabular}{p{0.22\linewidth}p{0.70\linewidth}}
\toprule
Metric & Definition \\
\midrule
PSNR$\uparrow$ & Peak Signal-to-Noise Ratio between aligned generated and reference RGB frames in $[0,1]$, averaged over time. \\
SSIM$\uparrow$ & Structural Similarity Index between aligned frames, computed with an $11\times11$ Gaussian window and averaged over time. \\
LPIPS$\downarrow$ & Learned Perceptual Image Patch Similarity using the AlexNet backbone, averaged over aligned frame pairs. \\
Kpt.Err$\downarrow$ & \textbf{Hand Keypoint End-Point Error}: evaluated on 50 clips selected from EgoDex. We compute the mean 2D Euclidean error between matched generated and reference hand keypoints, normalized by the image diagonal. We use ground-truth 2D joints when available; otherwise, the same MediaPipe 21-keypoint detector is applied to both generated and reference frames, followed by centroid-based hand matching. \\
Phys.Faith$\uparrow$ & Physical faithfulness, measured by the normalized single-axis score from Kimi K2.7 as defined below. It judges whether grasps, support, pushes, contact-driven object motion, non-penetration, and gravity are physically credible. \\
Subj. Cons.$\uparrow$ & \textbf{Subject Consistency}: VBench-style temporal consistency of DINO ViT-S/16 frame features, averaging cosine similarity to both the first and previous frames. \\
Bg. Cons.$\uparrow$ & \textbf{Background Consistency}: VBench-style temporal consistency of CLIP ViT-B/32 image features, using the same first-frame and adjacent-frame cosine-similarity aggregation. \\
\bottomrule
\end{tabular}
\end{table*}

\FloatBarrier

\subsection{Kimi K2.7 Prompt for Physical Faithfulness}

We use Kimi K2.7 as the judge. Frames are sampled uniformly from each generated video and presented in temporal order. The judge evaluates only whether contact and object dynamics are physically credible for the depicted manipulator type.

\begin{tcolorbox}[promptbox,title=Prompt: Physical Faithfulness]
\small
You are evaluating a generated video of manipulator--object interaction for the task ``[TASK PROMPT].'' You are shown $k$ frames sampled uniformly in temporal order. Treat a human hand, dexterous hand, gripper, or robot end effector according to its own embodiment.

Evaluate physical faithfulness considering whether:
\begin{enumerate}[leftmargin=1.5em,nosep]
    \item contacts, grasps, support, and pushes are credible;
    \item object motion is caused by plausible manipulator contact;
    \item objects avoid penetration, floating, and violations of gravity;
    \item contact and object dynamics remain coherent over time.
\end{enumerate}
Rate from 0--5, respond with only an integer.
\end{tcolorbox}

For the returned score $s_{\mathrm{phys}}\in\{0,\ldots,5\}$, we report
\begin{equation}
\operatorname{Phys.Faith}=\frac{s_{\mathrm{phys}}}{5}.
\end{equation}

\end{document}